\documentclass[preprint,11pt,authoryear,a4paper,nonatbib]{elsarticle}
\makeatletter
\let\c@author\relax
\makeatother

\usepackage{setspace}  % 加载setspace包用于设置行距
\usepackage{amssymb}
\usepackage{amsmath}
\journal{Transportation Research Part B}
\usepackage{hyperref}
\hypersetup{
colorlinks=true,
linkcolor=black,
citecolor=black,
urlcolor=black
}
\usepackage[style=apa,sorting=nyt,backend=biber]{biblatex}
\DeclareLanguageMapping{english}{english-apa}
\renewbibmacro{in:}{}
\DeclareFieldFormat{url}{\url{#1}}
\usepackage{enumitem}
\usepackage{float}
\usepackage{booktabs}
\usepackage{algorithm}
\usepackage{array}
\usepackage[left=20mm, right=20mm]{geometry}
\usepackage{caption}   % For table and figure captions
\usepackage[noend]{algpseudocode} % 使用 noend 选项隐藏结束语句
\usepackage{multirow}
\usepackage{graphicx}
\usepackage[table,xcdraw]{xcolor}
\usepackage[export]{adjustbox} % 用于图片垂直居中 valign=c
\graphicspath{{figs/}}

\begin{document}

\begin{frontmatter}



\title{Sequential Service Region Design with Capacity-Constrained Investment and Spillover Effect \tnoteref{t1}} %% Article title
\tnotetext[t1]{This research was supported by the National Natural Science Foundation of China (No.71971156 and No.72371188).}

\author[1,2]{Tingting Chen}

% \credit{Conceptualization of this study, Methodology, Software}

\affiliation[1]{organization={School of Economics and Management},
                addressline={Tongji University}, 
                city={Shanghai},
%               citysep={}, % Uncomment if no comma needed between city and postcode
                postcode={200092}, 
                country={China}}

\author[2]{Feng Chu}

\author[1]{Jiantong Zhang \corref{cor1}}
\cortext[cor1]{Corresponding author. Email: zhangjiantong@tongji.edu.cn.}

\affiliation[2]{organization={Laboratoire IBISC, Univ-Evry},
                addressline={Université Paris-Saclay}, 
                city={Evry},
                postcode={91000}, 
                % state={Orissa}, 
                country={France}}

%% Abstract
\begin{abstract}
Service region design determines the geographic coverage of service networks, shaping long-term operational performance. Capital and operational constraints preclude simultaneous large-scale deployment, requiring expansion to proceed sequentially. The resulting challenge is to determine when and where to invest under demand uncertainty, balancing intertemporal trade-offs between early and delayed investment and accounting for network effects whereby each deployment reshapes future demand through inter-regional connectivity. This study addresses a sequential service region design (SSRD) problem incorporating two practical yet underexplored factors: a $k$-region constraint that limits the number of regions investable per period and a stochastic spillover effect linking investment decisions to demand evolution. The resulting problem requires sequencing regional portfolios under uncertainty, leading to a combinatorial explosion in feasible investment sequences. To address this challenge, we propose a solution framework that integrates real options analysis (ROA) with a Transformer-based Proximal Policy Optimization (TPPO) algorithm. ROA evaluates the intertemporal option value of investment sequences, while TPPO learns sequential policies that directly generate high option-value sequences without exhaustive enumeration. Numerical experiments on realistic multi-region settings demonstrate that TPPO converges faster than benchmark DRL methods and consistently identifies sequences with superior option value. Case studies and sensitivity analyses further confirm robustness and provide insights on investment concurrency, regional prioritization, and the increasing benefits of adaptive expansion via our approach under stronger spillovers and dynamic market conditions.
\end{abstract}

\begin{highlights}
\item We study sequential service region design with k-region constraint and stochastic spillover effect
\item We formulate the problem as a Markov decision process under a real options framework to evaluate sequence-level option value
\item We develop a Transformer-based Proximal Policy Optimization algorithm to directly learn high option-value investment policies
\item Numerical experiments and case studies demonstrate superior option-value performance and robustness relative to benchmark strategies
\end{highlights}

%% Keywords

\end{frontmatter}



\section{Introduction}
\label{sec:1}
Service region design (SRD) addresses the strategic challenge of determining service area locations and coverage to meet customer demand \parencite{Huang_2018_Designing_S2,Yang_2024_Crowd_S5,Carlsson_2023_Provably_S7}. By shaping geographic coverage and resource allocation, SRD directly influences market penetration, operational efficiency, and long-term competitiveness in industries such as e-commerce and emerging transportation services. Although SRD has been extensively studied, much of the literature concentrates on continuous area partitioning or one-shot deployment decisions within a single planning horizon \parencite{WANG2017207_S1, Carlsson_2012_Dividing_S4, Huang_2018_Designing_S2, Banerjee_2022_Fleet_S6, Carlsson_2023_Provably_S7}. In contrast, we focus on discrete SRD, where investments are made over predefined geographic units \parencite{Grotschel}, reflecting administrative and operational constraints. In practice, many large-scale platforms engage in such SRD decisions. For example, \textcite{Amazon2024} regularly expands delivery zones to enhance coverage and fulfillment speed, while mobility-on-demand (MoD) platforms like Uber progressively invest in new cities and regions \parencite{Uber2025}. These SRD decisions require substantial capital and operational commitments and are undertaken under significant market uncertainty. As demand evolves alongside service rollout, firms must carefully determine not only where to deploy service coverage, but also how to prioritize geographic expansion in response to changing conditions.

Expanding service networks across multiple regions is a non-trivial problem. Prior SRD research predominantly adopts a one-shot framework, deciding service coverage for a single period and assuming that demand patterns remain stable rather than evolving over time \parencite{WANG2017207_S1, Banerjee_2022_Fleet_S6}. However,  full-scale, simultaneous investment across all regions is often risky and impractical due to the significant resources required. Consequently, service region expansion naturally gives rise to a sequential service region design (SSRD) problem, in which firms determine when and where to invest over time. Although phased deployment allows for risk mitigation and adaptation to evolving market conditions, identifying the optimal investment sequence requires careful evaluation of long-term returns: investing too early may lead to insufficient demand realization, whereas delaying expansion risks missed market opportunities. Existing SSRD studies predominantly formulate the problem as multi-period optimization models under deterministic or stationary stochastic demand, focusing on operational performance metrics such as cost, coverage, or service efficiency \parencite{Jafari_2017_Achieving_S9, Bender_2018_branch_S10, Liu_2025_Optimala_S11}, while several studies have utilized financial evaluation criteria such as net present value (NPV) under uncertainty \parencite{Goel_2004_stochastic_S13, Gupta_2017_Offshore_S14}. Nevertheless, NPV-based evaluation typically assumes fixed investment timing, without fully capturing managerial flexibility in adapting decisions as uncertainty unfolds. Real option analysis (ROA) provides a more appropriate framework by explicitly valuing such flexibility under uncertainty. Moreover, SSRD inherently exhibits nonstationary and state-dependent demand dynamics, as regional investments reshape future market conditions. This intertemporal feedback motivates a Markov decision process (MDP) formulation, yet its application to SSRD remains limited. To the best of our knowledge, \textcite{Rath_2024_deepa_S16} is among the few studies that model SSRD as an MDP and incorporate ROA to evaluate investment flexibility. They focused on MoD service deployment, utilizing supervised learning with recurrent neural networks (RNN) to predict the optimal investment sequence.

Our study builds on \textcite{Rath_2024_deepa_S16} by introducing additional structural features that more closely reflect real-world expansion dynamics. Specifically, we study a SSRD problem involving a set of geographically contiguous regions, where each region is invested exactly once within a finite planning horizon $\mathcal{N}$, where each region must be invested in exactly once within the planning horizon $\mathcal{T} = \{t_0, t_1, t_2, \ldots, t_T\}$, where $t_n$ is the $n$th time step in $\mathcal{T}$, ultimately forming a fully connected service network. Regions differ in size and exhibit nonstationary, uncertain demand that evolves over time, capturing customers’ dynamic market responses. Specifically, demand consists of intra-region demand, fulfilled once the region is invested, and inter-region (origin-destination) demand, fulfilled only when both connected regions are invested. The objective is to optimize the investment timing across regions to maximize returns. 

Importantly, we consider two practical yet underexplored factors. First, we impose a per-period investment capacity, allowing investment in at most $k$ regions at each time step. While prior studies (e.g., \textcite{Jafari_2017_Achieving_S9, Gupta_2017_Offshore_S14, Liu_2025_Optimala_S11}) have incorporated investment limitations through budget constraints, constraining the total expenditure. In contrast, the $k$-region constraint reflects realistic operational and resource limitations and fundamentally changes the decision structure: instead of sequencing individual regions, the planner must select portfolios of regions at each decision epoch. This transforms the problem from a pure permutation structure into a joint partition-and-sequencing problem, dramatically enlarging the combinatorial search space relative to single-region sequencing models \parencite{Rath_2024_deepa_S16} and introducing strong intertemporal dependencies across decisions. Second, we incorporate investment-induced spatial spillovers into demand evolution. Existing SSRD studies typically model nonstationary demand as an exogenous stochastic process \parencite{Gupta_2017_Offshore_S14, Ulmer_2022_Dynamic_S12, Rath_2024_deepa_S16}, without capturing cross-regional feedback generated by investment decisions. Although prior SRD research recognizes spatial interdependence in service coverage \parencite{HAN20201120_S3, Yang_2024_Crowd_S5, Qi_2018_Shareda_S8}, and economic studies document positive geographic spillovers in regional growth \parencite{Anousheh_2020_Spatial, DeBondt_1991_Strategic, Yu_2013_Spatial}, such mechanisms have rarely been integrated into sequential service region design. We therefore model both intra- and inter-regional demand as stochastically influenced by the spillover effect, capturing demand surges triggered by network expansion, enhanced connectivity, and cross-region adoption.

\begin{figure}[htbp]
    \centering
    \includegraphics[width=\textwidth]{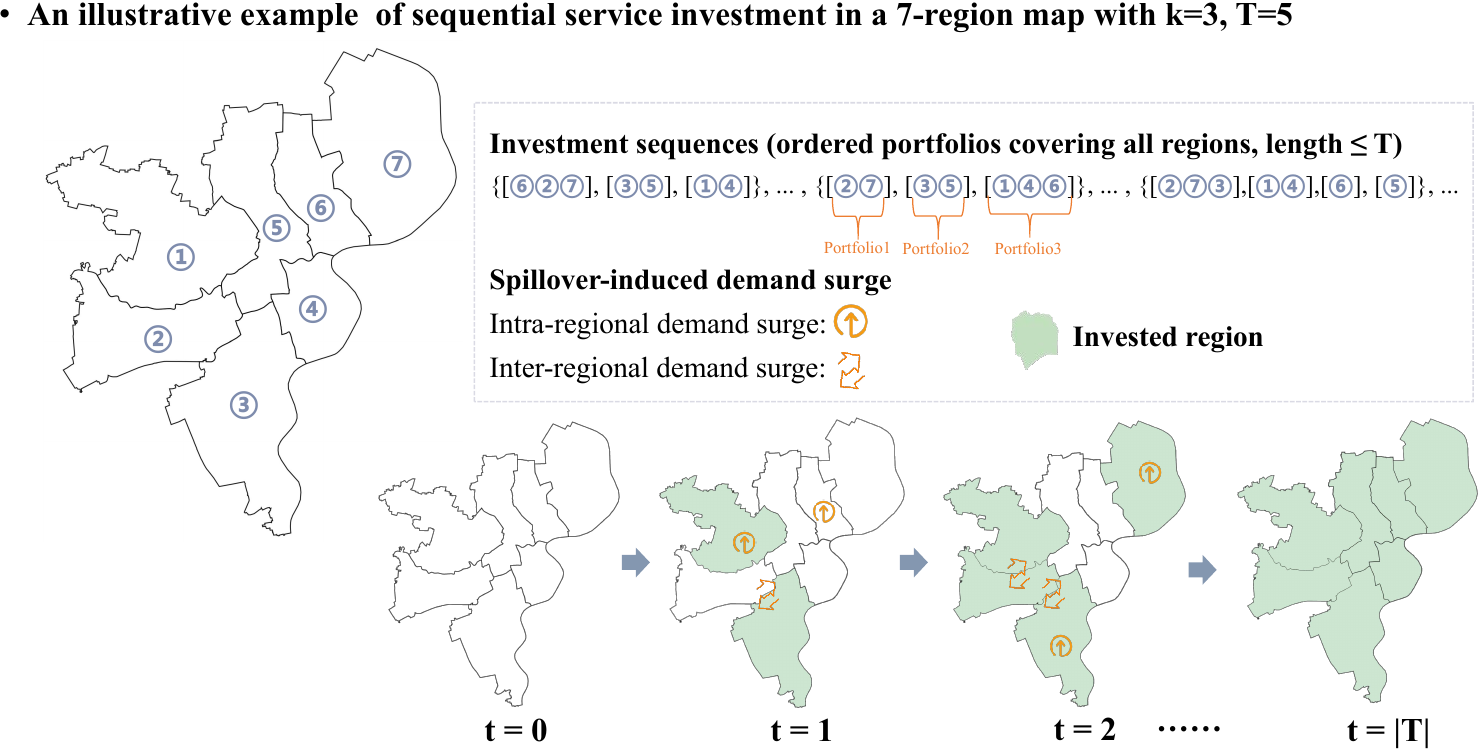}
    \caption{Illustration of the problem scenario}
    \label{fig:fig1}
\end{figure}

The computational challenge of our SSRD setting arises from the interaction between the k-region constraint and spillover-driven demand dynamics. As illustrated in Figure~\ref{fig:fig1}, consider a setting with $N=7$ regions over a planning horizon of $T=5$ periods and a portfolio size limit of $k=3$. At each decision epoch, the planner selects a subset of up to $k$ regions for simultaneous investment, while ensuring that each region is invested exactly once. All feasible investment sequences can be constructed by recursively enumerating candidate portfolios, i.e., subsets of size at most $k$, and forming ordered sequences of up to $T$ portfolios that collectively cover all regions exactly once. The number of possible portfolios is $m_p = \sum_{i=1}^{k} \binom{N}{i}$,and the resulting search space grows on the order of $O((m_p)^T)$. Even for moderate problem sizes, the number of feasible investment sequences increases rapidly. For instance, in the 7-region example with $k=3$, more than $25{,}000$ feasible sequences arise, and this number expands dramatically as $N$ or $k$ increases. Consequently, exhaustive enumeration and ROA-based valuation of all sequences quickly becomes computationally prohibitive. The challenge is further amplified by the endogenous spillover effect, as shown in Figure~\ref{fig:fig1}. Both intra- and inter-regional demand evolve stochastically and are influenced by ongoing investment decisions. Because current portfolio selections reshape future demand states, the value of any sequence depends on the evolving system trajectory. This state-dependent dynamic feedback motivates the development of adaptive solution approaches.

To address these challenges, we formulate SSRD as an MDP, employing ROA to evaluate the intertemporal option value of investment sequences. We develop a deep reinforcement learning (DRL) framework, termed Transformer-based Proximal Policy Optimization (TPPO), to directly learn high option-value investment policies. ROA provides sequence valuation feedback during training, while TPPO learns policies whose rollout generates complete investment sequences under the $k$-region constraint without exhaustive enumeration. From a theoretical perspective, this study advances the SSRD literature in three respects. First, we incorporate the $k$-region constraint, transforming the problem from sequencing individual regions to sequencing feasible regional portfolios. Second, we endogenize nonstationary demand evolution through the stochastic, investment-induced spillover effect, thereby capturing state-dependent network externalities. Third, we integrate ROA-based valuation with DRL-based policy learning, providing a scalable solution for high-dimensional SSRD problems. Extensive numerical experiments based on realistic multi-region settings, together with comprehensive sensitivity analyses, demonstrate the effectiveness and robustness of TPPO. The results yield several managerial insights. First, optimal expansion strategies exhibit structural stability across different $k$-region constraints, and moderate investment concurrency is observed to deliver superior real option value under increasing market growth and volatility. Second, the learned sequences follow a bottom-up expansion logic, where smaller or lower baseline demand regions are deployed earlier, while larger or higher baseline demand regions are strategically deferred. Third, high-value sequences further indicate that certain regions are more effectively deployed in combination, suggesting structured complementarities in regional investment. Finally, the performance advantage of TPPO becomes more pronounced as spillover intensity strengthens, the scale effect increases, or service costs decline over time, highlighting the value of adaptive investment policies in dynamic environments.

The remainder of this paper is structured as follows. Section \ref{sec:2} reviews the related literature. Section \ref{sec:3} presents the model formulation. Section \ref{sec:4} introduces the proposed solution approach. The results of extensive computational experiments are presented in Section \ref{sec:5}. Finally, Section \ref{sec:6} contains concluding remarks and managerial insights.

\section{Literature review}
\label{sec:2}
In this section, we review the literature related to our study: service region design (SRD) and real option analysis (ROA)

\subsection{Service region design (SRD)}
\label{sec:2.1}
Service region design (SRD) concerns the strategic determination of geographic service coverage. We focus on discrete SRD studies, where regions become discrete decision units whose spatial interdependence and heterogeneous demand profiles give rise to a combinatorial investment problem. Current studies can also be differentiated by planning mode, including one-shot planning and multi-step planning. Most studies adopt a one-shot framework, particularly in urban delivery contexts, where service regions are determined for a single planning horizon and integrated with routing or facility allocation decisions. 
These formulations typically model SRD as a static optimization problem under deterministic demand assumptions \parencite{WANG2017207_S1, Huang_2018_Designing_S2, Carlsson_2012_Dividing_S4}, or extend to stationary stochastic settings where demand follows stable distributions and region configurations remain fixed over time \parencite{Banerjee_2022_Fleet_S6, Carlsson_2023_Provably_S7, Qi_2018_Shareda_S8, Yang_2024_Crowd_S5}.

Determining service regions only once may be inadequate in environments where demand evolves over time and investment decisions generate long-term operational consequences. Accordingly, several studies extend SRD to sequential SRD (SSRD), formulating multi-period planning models that optimize regional configurations under both deterministic \parencite{Bender_2018_branch_S10} and stochastic demand conditions \parencite{Goel_2004_stochastic_S13, Gupta_2017_Offshore_S14, Ulmer_2022_Dynamic_S12}. In deterministic SSRD settings, the entire planning horizon is typically formulated as a single integrated optimization problem \parencite{Bender_2018_branch_S10}, where region assignments are jointly determined across periods. When uncertainty is incorporated, different stochastic modeling frameworks have been adopted. \textcite{Goel_2004_stochastic_S13} and \textcite{Gupta_2017_Offshore_S14} formulated investment planning problems as multistage stochastic programs, in which uncertainty unfolds over discretized scenario trees but the underlying stochastic structure does not explicitly model time-varying distributional parameters. In contrast, \textcite{Ulmer_2022_Dynamic_S12} modeled dynamic service area sizing as a stochastic dynamic decision process with time-dependent arrival rates, thereby allowing for nonstationary stochastic demand. Meanwhile, \textcite{Lei_2015_Dynamic_S15} considered nonstationary demand in a deterministic multi-period districting framework, where demand evolves over time through exogenously specified updates.

SSRD inherently exhibits state-dependent dynamics, as regional investments alter future demand conditions under nonstationary uncertainty. 
An MDP formulation therefore provides a natural framework, representing investment decisions as policies over system states with explicit transition dynamics.  However, MDP-based approaches to sequential SRD remain limited. To our knowledge, \textcite{Rath_2024_deepa_S16} is among the few studies that model service zone expansion as an MDP. They represented nonstationary demand using Geometric Brownian Motion (GBM), evaluated investment flexibility via real option analysis (ROA), and employed recurrent neural networks (RNNs) to identify high-value investment sequences. 
Similarly, our study adopts an ROA-based evaluation framework for sequential regional investments under nonstationary demand uncertainty. In contrast, we extend this line of work in two important aspects. 
First, we introduce a per-period investment capacity constraint that limits the number of regions selected at each decision epoch. While prior studies (e.g., \textcite{Jafari_2017_Achieving_S9, Gupta_2017_Offshore_S14, Liu_2025_Optimala_S11}) incorporate investment limitations through per-period budget constraints, such formulations regulate investment intensity in a continuous manner and do not explicitly control the number of simultaneous investment actions. In contrast, the proposed $k$-region constraint transforms the decision process into a combinatorial portfolio selection problem, where decisions are made over subsets of regions and early selections restrict the feasible set of future investment portfolios. Second, we incorporate the spatial spillover effect into demand evolution. While GBM captures temporal demand uncertainty (e.g., \textcite{Rath_2024_deepa_S16}), it neglects cross-regional interactions induced by investment decisions. Empirical evidence in regional economics and transportation shows that infrastructure investments generate positive spillovers across regions through enhanced connectivity and market integration \parencite{Pradhan_2013_Effect, Arbues_2015_spatial, Qi_2020_Spatial}. 
To capture this mechanism, we extend the GBM process by introducing Poisson jump components, where jumps represent discrete, investment-induced increases in regional demand.

\subsection{Real option analysis (ROA)}
\label{sec:2.2}
Existing SRD studies primarily evaluate solutions using operational metrics such as cost, service coverage, equity, and service time, which are appropriate for one-shot planning settings \parencite{Carlsson_2012_Dividing_S4, WANG2017207_S1, Huang_2018_Designing_S2, Banerjee_2022_Fleet_S6}. In multi-step contexts, however, regional expansion becomes a strategic investment decision. Financial measures, most notably net present value (NPV), assess project viability by discounting expected future cash flows \parencite{Goel_2004_stochastic_S13, Gupta_2017_Offshore_S14}. Nevertheless, NPV assumes fixed investment timing and does not capture managerial flexibility in staging, delaying, or adapting decisions as uncertainty evolves. In sequential SRD, where expansion can be exercised progressively, such flexibility constitutes a significant source of value.
Real Option Analysis (ROA) extends the NPV framework by valuing the option to defer, expand, contract, or abandon investments in response to stochastic demand dynamics \parencite{McGrath_1997_Real}. By embedding investment timing within a stochastic process, ROA captures both downside risk and upside potential, making it particularly suitable for sequential regional expansion under nonstationary uncertainty.

ROA has been widely applied across diverse investment domains. For example, in information technology (IT) strategy, ROA has been used to structure investment decisions by identifying core components of IT value and strategic actions \parencite{Kim_2002_Strategic_R1}. \textcite{Duku-Kaakyire_2004_R2} applied ROA in forestry investment to evaluate four management options: delaying reforestation, expanding processing capacity, abandoning the plant or takeover, and a combined assessment of these options. Research on ROA in transportation networks emerges as a prominent area of study. \textcite{Chow_2011_Real_R6} reviewed network design models and proposed an analytical method to quantify the value of flexibility in deferring and redesigning network investments under nonstationary stochastic demand using ROA techniques. \textcite{Chow_2011_Network_R7} adopted a real options framework in network design by maximizing expanded net present value (ENPV), which augments traditional NPV with option value arising from investment deferral and interaction effects. A comprehensive review of ROA applications in transportation, including real option game models and multimodal infrastructure investments, is provided by \textcite{Trigeorgis_2017_R8}.

ROA solution approaches generally fall into three categories: finite difference methods, lattice models, and simulation-based techniques \parencite{Chow_2011_Real_R6}. Finite difference and lattice methods discretize the underlying stochastic process but become computationally challenging in high-dimensional settings due to the curse of dimensionality. As a result, simulation-based approaches—particularly Monte Carlo methods—have become widely adopted for valuing complex real options. To address the limitation of traditional forward-looking Monte Carlo methods in handling American-style options, \textcite{Longstaff_2001_Valuing} developed the Least Squares Monte Carlo (LSM) method, which combines simulation with regression to approximate optimal stopping policies. LSM is especially effective for high-dimensional and path-dependent problems and has been extended to settings with multiple interacting real options \parencite{Gamba_2003_Real_R3, Gamba_2002_Extension_R4}. Such simulation-based ROA methods have been widely applied in network investment and infrastructure planning contexts \parencite{Chow_2011_Network_R7, Trigeorgis_2017_R8}. Table~\ref{tab:table1} summarizes the related works and highlights the key distinctions between our research and prior work. 

\begin{table}[htbp]
\resizebox{\textwidth}{!}{%
\begin{tabular}{@{}
>{\columncolor[HTML]{FFFFFF}}c 
>{\columncolor[HTML]{FFFFFF}}c 
>{\columncolor[HTML]{FFFFFF}}c 
>{\columncolor[HTML]{FFFFFF}}c 
>{\columncolor[HTML]{FFFFFF}}c 
>{\columncolor[HTML]{FFFFFF}}c 
>{\columncolor[HTML]{FFFFFF}}c 
>{\columncolor[HTML]{FFFFFF}}c 
>{\columncolor[HTML]{FFFFFF}}c 
>{\columncolor[HTML]{FFFFFF}}c @{}}
\toprule
\cellcolor[HTML]{FFFFFF} & \cellcolor[HTML]{FFFFFF} & \multicolumn{3}{c}{\cellcolor[HTML]{FFFFFF}Demand} & \cellcolor[HTML]{FFFFFF} & \cellcolor[HTML]{FFFFFF} & \cellcolor[HTML]{FFFFFF} & \cellcolor[HTML]{FFFFFF} & \cellcolor[HTML]{FFFFFF} \\ \cmidrule(lr){3-5}
\cellcolor[HTML]{FFFFFF} & \cellcolor[HTML]{FFFFFF} & \cellcolor[HTML]{FFFFFF} & \multicolumn{2}{c}{\cellcolor[HTML]{FFFFFF}Uncertain} & \cellcolor[HTML]{FFFFFF} & \cellcolor[HTML]{FFFFFF} & \cellcolor[HTML]{FFFFFF} & \cellcolor[HTML]{FFFFFF} & \cellcolor[HTML]{FFFFFF} \\ \cmidrule(lr){4-5}
\multirow{-3}{*}{\cellcolor[HTML]{FFFFFF}Reference} & \multirow{-3}{*}{\cellcolor[HTML]{FFFFFF}Type of planning} & \multirow{-2}{*}{\cellcolor[HTML]{FFFFFF}Deterministic} & Stationary & Nonstationary & \multirow{-3}{*}{\cellcolor[HTML]{FFFFFF}Investment limitation} & \multirow{-3}{*}{\cellcolor[HTML]{FFFFFF}Uncertain spillover effect} & \multirow{-3}{*}{\cellcolor[HTML]{FFFFFF}Evaluation indicator} & \multirow{-3}{*}{\cellcolor[HTML]{FFFFFF}Modeling} & \multirow{-3}{*}{\cellcolor[HTML]{FFFFFF}Method} \\ \midrule
Carlsson et al. (2013) & \cellcolor[HTML]{FFFFFF} & \checkmark &  &  &  &  & SE & MINLP & WVD \\
Wang and Lin (2017) & \cellcolor[HTML]{FFFFFF} & \checkmark &  &  &  &  & C & MILP & CSA \\
Huang et al. (2018) & \cellcolor[HTML]{FFFFFF} & \checkmark &  &  &  &  & C & MILP & ALNS \\
Qi et al. (2018) & \cellcolor[HTML]{FFFFFF} &  & \checkmark &  &  &  & C & CA & NA \\
Banerjee et al. (2022) & \cellcolor[HTML]{FFFFFF} &  & \checkmark &  &  &  & SC & CA & WCVT \\
Yang et al. (2024) & \cellcolor[HTML]{FFFFFF} &  & \checkmark &  &  &  & SE & MINLP & ACO+TL \\
Carlsson et al. (2024) & \multirow{-7}{*}{\cellcolor[HTML]{FFFFFF}One-shot} &  & \checkmark &  &  &  & ST & CA+Queuing & RPA+NA \\ \midrule
Jafari et al. (2017) & \cellcolor[HTML]{FFFFFF} & \checkmark &  &  & \textbf{\checkmark} &  & C & MIP & HM \\
Bender (2018) & \cellcolor[HTML]{FFFFFF} & \checkmark &  &  &  &  & ST & MILP & B\&P \\
Liu et al. (2025) & \cellcolor[HTML]{FFFFFF} & \checkmark &  &  & \textbf{\checkmark} &  & C & MILP & HM \\
Goel et al. (2004) & \cellcolor[HTML]{FFFFFF} &  & \checkmark &  &  &  & NPV & MILP & AM \\
Gupta et al. (2017) & \cellcolor[HTML]{FFFFFF} &  & \checkmark &  & \checkmark &  & NPV & MINLP & LD \\
Lei et al. (2015) & \cellcolor[HTML]{FFFFFF} &  &  & \checkmark &  &  & C & MILP & ALNS \\
Ulmer et al. (2022) & \cellcolor[HTML]{FFFFFF} &  &  & \checkmark &  &  & SC & CA & VFA \\
Rath and Chow (2024) & \cellcolor[HTML]{FFFFFF} &  &  & \checkmark &  &  & ROV & MDP + GBM & ROA+RNN \\
\textbf{This work} & \multirow{-9}{*}{\cellcolor[HTML]{FFFFFF}Multi-step} & \textbf{} & \textbf{} & \textbf{\checkmark} & \textbf{\checkmark} & \textbf{\checkmark} & \textbf{ROV} & \textbf{MDP + GBMPJ} & \textbf{ROA+TPPO} \\ \bottomrule
\end{tabular}%
}
\vspace{-2pt}
\scriptsize
\textit{Note:} SE: service equity; C: cost; SC: service coverage; ST: service time; NPV: net present value; ROV: real option value; MINLP: mixed integer non linear programming; MILP: mixed integer linear programming; CA: continuous analysis; MDP: markov decision process; GBM: genometric brownian motion; GBMPJ: GBM with EPoisson jump; WVD: weighted Voronoi diagram; CSA: clustering scenario-based algorithm; ALNS: adaptive large neighborhood search; NA: numerical analysis; WCVT: weighted centroidal Voronoi tessellation; ACO: ant colony optimization; TL: transfer learning; RPA: recursive partitioning algorithm; HM: heuristic method; B\&P: branch\&price; AM: approximation method; LD: lagrangian decomposition; VFA: value function approximation; ROA: real option analysis; RNN: recurrent neural network; TPPO: Transformer-based proximal policy approximation
\caption{Comparison with related works}
\label{tab:table1}
\end{table}

\section{Model formulation}
\label{sec:3}

\subsection{Nonstationary uncertainty demand under spillover effect}
\label{sec:3.1}
We formally define the problem as follows. Recalling that $\mathcal{T} = \{t_0, t_1, t_2, \ldots, t_T\}$ is a finite set of discrete time periods within the planning horizon. A service operator seeks to deploy services across $N$ geographically contiguous regions, which collectively form a fully connected service network represented by a graph $G = (\mathcal{N}, \mathcal{E})$, where $\mathcal{N}$ denotes the set of regions and $\mathcal{E}$ the set of links connecting each pair of regions. Each region is to be invested in exactly once during $\mathcal{T}$. Upon investment, a region must serve both intra-region demand—satisfied locally—and inter-region (origin-destination, OD) demand, which arises between any two regions only after both have been invested in. Specifically, let $Q_{ijt_n} \in \mathbb{R}^{|\mathcal{N}| \times |\mathcal{N}|}$ represent the OD demand from region $i$ to region $j$ at time $t_n$, with $i, j \in \mathcal{N}$. For notational convenience, $Q_{iit_n}$ denotes the intra-region demand for region $i$ at time $t_n$. This modeling framework reflects the structure of real-world service networks, such as those found in Mobility-on-Demand and ridesharing systems.

To model nonstationary uncertainty in origin-destination (OD) service demand, the Geometric Brownian Motion (GBM) process is widely utilized. GBM is a continuous-time stochastic process where the logarithm of the variable follows a Brownian motion, and has been extensively used to model travel and vehicular demand in service networks \parencite{Gao_2013_Incorporating, Li_2015_Transita}. Consistent with prior work on service region design \parencite{Chow_2011_Network_R7}, we assume independence between regions and model the demand evolution for each OD pair $ij$ as:

\begin{equation}  
    \frac{dQ_{ijt}}{Q_{ijt}} = \mu dt + \sigma dW_{t} \label{eq:gbm}  
\end{equation}
where demand evolves continuously over time $t\in[0,T]$ and $dt$ is an infinitesimal time increment, $\mu$ represents the drift parameter, $\sigma$ denotes the volatility rate, and $dW_t \sim N(0, dt)$ is a standard Wiener process.

While standard GBM captures temporal demand evolution, we incorporate a stochastic spillover effect, which dynamically amplifies both intra- and inter-region demand during the investment process. Specifically, we extend the GBM model by considering a Poisson jump component (GBMPJ), enabling the representation of sudden demand surges analogous to those observed in financial markets. The Poisson jump process, widely utilized in financial modeling for risk management and option pricing, combines a continuous diffusion term (modeled by Brownian motion) with a jump term (modeled by a Poisson process) \parencite{Yan2025BayesianIO}. In our context, the OD demand for each region is analogous to a market price, where the spillover effect induces abrupt demand jumps. Let $\eta$ denote a stochastic random variable representing the uncertain spillover effect, and let $f(I_t)$ be a monotonically increasing function of the cumulative number of invested regions $I_t$ up to time $t$. The service demand at each OD pair $ij$ is then modeled as: 

\begin{equation}  
\frac{dQ_{ijt}}{Q_{ijt}} = \mu dt + \sigma dW_{t} + \eta \cdot f(I_t) dJ_{it}  
\end{equation}
where $J_{it}$ denotes a region-specific Poisson process with intensity $\lambda_i$. The intensity parameter $\lambda_i$ is not exogenously fixed; instead, it is systematically determined by the interaction between normalized population density and spatial coverage indices, capturing heterogeneous spillover potential across regions. The diffusion and jump components are assumed mutually independent and independent of $\eta$.Following \textcite{Rath_2024_deepa_S16}, we allow region-specific volatility $\sigma \in {\sigma_1, \sigma_2, \ldots, \sigma_{|\mathcal{N}|}}$, where all OD pairs originating from region $i$ share volatility parameter $\sigma_i$.

Furthermore, we impose a constraint limiting investments to at most $k$ regions per time step. Let $\mathcal{L}$ denote the set of all feasible investment sequences and define $z$ as a portfolio comprising a subset of regions from $\mathcal{N}$ with $|z| \leq k$. Each $l = \{z_1, z_2, \ldots, z_H\} \in \mathcal{L}$ represents an ordered sequence of portfolios that collectively covers all regions in $\mathcal{N}$ exactly once. The objective is to identify the optimal investment sequence that maximizes investment return over the planning horizon. Intuitively, the investment sequence can be constructed incrementally through the selection of portfolios over time. This involves determining the optimal sequence of portfolios, where each portfolio is formed from the set of uninvested regions at each time step, based on available information and prior investment decisions. However, the number of candidate portfolios at each time step can be extremely large and grows exponentially with $k$. Moreover, allowing the option to skip investments in any given period significantly increases the number of possible investment paths. In Section~3.2, we detail the generation of potential investment sequences and introduce a real options analysis (ROA) model designed to evaluate these sequences and identify the optimal investment solution efficiently.
    
\subsection{ROA for sequence evaluation}
\label{sec:3.2}

This section introduces ROA for evaluating investment sequences. 
Prior to valuation, it is necessary to construct a feasible investment sequence. 
We assume that any unused budget remains available throughout the planning horizon and can be allocated to future investments. 
To accommodate the timing flexibility of portfolio investing, we initially focus on the relative ordering of portfolios within an investment sequence rather than explicit investment timing. 
An investment sequence is deemed feasible if it satisfies three conditions: 
(1) its length does not exceed the planning horizon $\mathcal{T}$; 
(2) each portfolio contains at most $k$ regions; and 
(3) every region is included exactly once across all portfolios in the sequence. 
The ordered portfolios thus specify an investment order over the planning horizon. 
We denote the set of all feasible investment sequences by $\mathcal{L}$, where a sequence is written as $l=(z_1,z_2,\ldots,z_H)\in\mathcal{L}$ with each portfolio $z_h\subseteq \mathcal{N}$.

The challenge then becomes determining the explicit investment timing for portfolios within a given sequence, subject to the constraint that at most one portfolio can be invested per time step. 
Following Dixit and Pindyck (1994), the problem of deciding the optimal time to exercise an option can be interpreted as an optimal stopping problem, i.e., a dynamic programming control problem that optimizes a binary decision to continue (defer) or stop (invest) at each time step \parencite{Chow_2011_Real_R6}. 
To achieve this, we propose a ROA evaluation procedure based on the Least Squares Monte Carlo (LSMC) method \parencite{Gamba_2003_Real_R3} to evaluate the policy value (i.e., option value) of any given sequence $l\in\mathcal{L}$. 
Specifically, the method integrates Monte Carlo (MC) simulations with a forward simulation of uncertainty and a backward recursion (least-squares regression) to estimate continuation values. 
Our objective is to determine the investment/deferral decision $a_{z_h,t_n}\in\{0,1\}$ for each portfolio $z_h$ at each time step $t_n$, where $a_{z_h,t_n}=1$ indicates investing immediately and $a_{z_h,t_n}=0$ indicates deferring. 
The sequence exhibits compound option characteristics, where the value of subsequent options is incorporated into the continuation value when evaluating earlier portfolios \parencite{Gamba_2003_Real_R3}.

\paragraph{\textbf{Bellman recursion and immediate payoff}}
This framework yields the following Bellman equation for determining the option value to invest in portfolio $z_h$ as a function of stochastic OD demand:
\begin{align}
F_{z_h}(t_n, \mathbf{X}_{z_h, t_n}) = \max \Bigg\{\, 
    & \pi_{z_h}(t_n, \mathbf{X}_{z_h, t_n}) + (1+\rho)^{-1} \mathbb{E}_{t_n} \left[ F_{z_{h+1}}(t_{n+1}, \mathbf{X}_{z_{h+1}, t_{n+1}}) \right] \notag \\
    &,\ (1+\rho)^{-1} \mathbb{E}_{t_n} \left[ F_{z_h}(t_{n+1}, \mathbf{X}_{z_h, t_{n+1}}) \right] 
\Bigg\},
\end{align}
where $\mathbf{X}_{z_h, t_n}$ is the state variable associated with portfolio $z_h$, representing the estimated incremental demand brought by adding $z_h$ to the current service network at time $t_n$, which is a function of the stochastic OD demand generated from the regions in the portfolio at time $t_n$. 
Specifically, $\mathbf{X}_{z_h, t_n} = \mathbf{X}^{sum}_{z_h, t_n} - \mathbf{X}^{sum}_{z_{h-1}, t_n}$, where $\mathbf{X}^{sum}_{z_h, t_n}$ is the total demand summing both intra- and inter-region demand of the regions included in portfolios $z_1,\ldots,z_h$ at $t_n$, and $\mathbf{X}^{sum}_{z_{h-1}, t_n}$ is defined analogously for portfolios $z_1,\ldots,z_{h-1}$. 
Here, $\rho$ is the risk-free discount rate, and $\pi_{z_h}(t_n, \mathbf{X}_{z_h, t_n})$ denotes the present worth of an immediate investment in portfolio $z_h$ at time $t_n$.

We assume that the payoff from investing in portfolio $z_h$ is given by the incremental demand realized from including $z_h$ in the service network at time $t_n$ (i.e., $\mathbf{X}_{z_h,t_n}$), net of a service threshold. 
This threshold is analogous to the strike price in financial options \parencite{Longstaff_2001_Valuing} and is expressed as $|z_h|c_{intra} + \frac{|z_h|(2|\mathcal{N}_{z_h}|-|z_h|-1)}{2}c_{inter}$, where $c_{intra}$ denotes the intra-region cost incurred for serving local demands and $c_{inter}$ denotes the inter-region cost incurred for serving new connections. 
Accordingly, the immediate payoff can be written as
\[
\pi_{z_h}(t_n, \mathbf{X}_{z_h,t_n})
=
\mathbf{X}_{z_h,t_n}
-
\left(
|z_h|\,c_{intra}
+
\frac{|z_h|\big(2|\mathcal{N}_{z_h}|-|z_h|-1\big)}{2}
\,c_{inter}
\right).
\]

The inter-region component $\frac{|z_h|(2|\mathcal{N}_{z_h}|-|z_h|-1)}{2}c_{inter}$ represents the cost associated with the new inter-region links created when portfolio $z_h$ is added. 
Here, $|z_h|$ denotes the cardinality of $z_h$, and $|\mathcal{N}_{z_h}|$ denotes the total number of regions in the service network after including $z_h$. 
In particular, adding $z_h$ generates $\frac{|z_h|(2|\mathcal{N}_{z_h}|-|z_h|-1)}{2}$ new inter-region connections with the existing $h-1$ portfolios in the investment sequence.
For illustration, consider a sequence $\{z_1,z_2\}$ where $z_1=\{r_1,r_2,r_3\}$ and $z_2=\{r_4,r_5\}$. 
Starting with $z_1$, the service network contains three regions ($r_1,r_2,r_3$) and three inter-region links ($r_1$--$r_2$, $r_1$--$r_3$, $r_2$--$r_3$), yielding a cost of $3c_{intra}+3c_{inter}$. 
After adding $z_2$, with $|z_2|=2$ and $|\mathcal{N}_{z_2}|=5$, seven additional inter-region links are created, so the incremental cost becomes $2c_{intra}+7c_{inter}$.

\paragraph{\textbf{LSMC estimation procedure}}
The total payoff for portfolio $z_h$ consists of the immediate payoff from investing in $z_h$ plus the discounted expected option value of investing in the subsequent portfolio $z_{h+1}$ at the next time step, given by $(1+\rho)^{-1}\mathbb{E}_{t_n}\!\left[F_{z_{h+1}}(t_{n+1},\mathbf{X}_{z_{h+1},t_{n+1}})\right]$. 
This quantity is compared with the continuation value of deferring investment in $z_h$, defined as $(1+\rho)^{-1}\mathbb{E}_{t_n}\!\left[F_{z_h}(t_{n+1},\mathbf{X}_{z_h,t_{n+1}})\right]$. 
We denote the continuation value for $z_h$ by $\phi_{z_h}(t_n,\mathbf{X}_{z_h,t_n})$. 
Both conditional expectations are estimated using LSMC regression, which fits the continuation value as a linear combination of basis functions of the current state variables.

Specifically, for each $t_n\in\mathcal{T}$, the $|\mathcal{N}|\times|\mathcal{N}|$ OD demand matrix is simulated across a set of Monte Carlo paths $\Omega$ using the GBMPJ process, and $\phi_{z_h}(t_n,\mathbf{X}_{z_h,t_n})$ is estimated for each realization $\omega\in\Omega$. 
To estimate $\phi_{z_h}(t_n,\mathbf{X}_{z_h,t_n}(\omega))$, which represents the value of deferring investment in $z_h$ to await future OD demand realizations along path $\omega$, we compare this continuation value with the immediate payoff to determine the optimal stopping time. 
Let $\tau_{z_h}(\omega)$ denote the optimal stopping time for portfolio $z_h$ along path $\omega$. 
Along each path, if the immediate payoff exceeds the continuation value at $t_n$, the investment is exercised and the stopping time is updated to $t_n$. 
The stopping times must satisfy the sequential constraint $\tau_{z_h}(\omega)\ge \tau_{z_{h-1}}(\omega)+1$ for $h>1$ to maintain the ordered investment sequence. 
This optimization is accomplished through a backward recursion from maturity $t_T$ to $t_0$ and from portfolio $z_H$ to $z_1$, which determines the policy value of the investment sequence (see Section~4 for implementation details).

We employ Hermite polynomials as orthonormal basis functions, where $L_j$ represents the $j$-th basis function and $U$ denotes the total number of basis functions. 
The continuation value $\phi_{z_h}(t_n,\mathbf{X}_{z_h,t_n}(\omega))$ for path $\omega$ is approximated as
\begin{align*}
\phi_{z_h}(t_n,\mathbf{X}_{z_h,t_n}(\omega))
&\approx
\phi^{*}_{z_h,t_n}(\omega)
=
\sum_{j=1}^{U}\beta^{*}_{j}(t_n)\,L_j(\mathbf{X}_{z_h,t_n}(\omega)),
\end{align*}
where $\beta^{*}_{j}(t_n)$ are the regression coefficients obtained via least-squares estimation across all simulation paths. 
Similarly, $\phi_{z_{h+1}}(t_n,\mathbf{X}_{z_{h+1},t_n}(\omega))$ is estimated using the same LSMC approach with basis functions applied to the state variables of portfolio $z_{h+1}$ at time $t_n$.

Therefore, for any investment sequence $l=\{z_1,\dots,z_H\}\in\mathcal{L}$, we denote by $V_{\mathrm{ROA}}(l)$ the optimal option value computed via the above ROA evaluation. 
This value represents the discounted expected payoff under the optimal stopping rule along sequence $l$.

\section{Solution approach}
\label{sec:4}

\subsection{Solution algorithm for ROA}
\label{sec:4.1}

% 定义输入输出标签
\algrenewcommand{\algorithmicrequire}{\textbf{Input:}}
\algrenewcommand{\algorithmicensure}{\textbf{Output:}}

% 优化伪代码结构（减少缩进）
\algrenewcommand{\algorithmicindent}{0.5em} % 调整缩进量
\renewcommand{\baselinestretch}{0.9} % 全局行距调整
\begin{algorithm}[H]
\scriptsize % 设置算法字体大小
\caption{Real option analysis (ROA) algorithm for sequence evaluation} % 标题必须在算法环境内顶部
\label{alg:ROA} % 标签紧跟标题后

\begin{algorithmic}[1]
    \Require 
      Investment sequence $l = \{z_1,z_2,\dots,z_H\}$, 
      Time steps $\mathcal{T} = \{t_0,t_1,\dots,t_T\}$, 
      Stopping times $\tau_{z_h} \gets t_T$, Real option values $F_{z_h}$, Immediate payoffs $\pi_{z_h}$, State variables $X_{z_h,t_n}\ \forall z_h \in s, \forall t_n \in \mathcal{T}$,  
     Spillover effect $\eta$,
      Intra-/Inter-regional costs $c_{intra}, c_{inter}$,
      Number of Monte Carlo paths $|\Omega|$,
      Discount rate $\rho$
    \Ensure 
      Option value of the sequence $V_{\mathrm{ROA}}(l)$,
    
    \State \textbf{Terminal condition setup:}
    \For{$h = 1$ \textbf{to} $H$}
        \For{$\omega = 1$ \textbf{to} $|\Omega|$}
            \State $t_{e} \gets T - (H - h) + 1$
            \State $\tau_{z_h}(\omega) \gets t_{e}$
            \State $\pi_{z_h}(t_{e},X_{z_h,t_{e}}(\omega)) \gets \text{Immediate\_Payoff}(c_{ir}, c_{wr}, t_{e}, X_{z_h,t_{e}}(\omega), \eta_)$
            \State $F_{z_h}(t_{e},X_{z_h,t_{e}}(\omega)) \gets \pi_{z_h}(t_{e},X_{z_h,t_{e}}(\omega))$
        \EndFor
    \EndFor
    
    \State \textbf{Dual backward recursion:}
    \For{$t_n = t_T-1$ \textbf{to} $t_0$} 
        \For{$h = H$ \textbf{downto} $1$}
            \If{$t_n <= \tau_{z_h}(\omega)$}
                \State Generate $|\Omega|$ Monte Carlo paths $\Omega_{z_h,t_n}$
                \State Compute $\pi_{z_h}(t_n,X_{z_h,t_n}(\omega)) \gets \text{Immediate\_Payoff}(c_{ir}, c_{wr}, X_{z_h,t_n}(\omega), \eta)$\ $\forall \omega \in \Omega_{z_h,t_n}$ 
            \EndIf
            
            \If{$h < H-1$}
                \State Calculate the expected discounted value of the next portfolio: 
                \State $\phi_{z_{h+1}}(t_n, X_{z_{h+1},t_n}(\omega)) \gets E_{t_n}\left((1+\rho)^{(t_{n+1}-t_n)}F_{z_{h+1}}(t_{n+1},X_{z_h,t_n}(\omega))\right)\ \forall \omega \in \Omega_{z_{h+1},t_{n+1}}$ (LSMC estimation)
                \State Calculate continuation value: 
                \State $\phi_{z_{h}}(t_n, X_{z_{h},t_n}(\omega)) \gets E_{t_n}\left((1+\rho)^{(t_{n+1}-t_n)}F_{z_h}(t_{n+1},X_{z_h,t_{n+1}}(\omega))\right)\ \forall \omega \in \Omega_{z_h,t_{n+1}}$ (LSMC estimation)
            \Else
                \State $\varphi_{z_{h+1},t_{n+1}} \gets 0$; $\phi_{z_h,t_{n+1}} \gets 0$
            \EndIf
            
            \For{$\omega = 1$ \textbf{to} $|\Omega|$}
                \If{$t_n <= \tau_{z_h}(\omega)$}
                    \If{$\pi_{z_h}(t_n,X_{z_h,t_n}(\omega)) + \varphi_{z_{h+1},t_{n+1}} \geq \phi_{z_h,t_{n+1}}$}
                        \State $\tau_{z_h}(\omega) \gets t_n$
                        \State $F_{z_h}(t_n,X_{z_h,t_n}(\omega)) \gets \pi_{z_h}(t_n,X_{z_h,t_n}(\omega)) + \varphi_{z_{h+1},t_{n+1}}$
                        \For{$m = h+1$ \textbf{to} $H$}
                            \State $\tau_{z_m}(\omega) \gets \max(\tau_{z_m}(\omega), \tau_{z_{m-1}}(\omega)+1)$
                        \EndFor
                    \Else
                        \For{$m = h$ \textbf{to} $H$}
                            \State $F_{z_m}(t_n,X_{z_m,t_n}(\omega)) \gets (1+\rho)^{(t_{n+1}-t_n)} \cdot F_{z_m}(t_{n+1},X_{z_m,t_n}(\omega))$
                        \EndFor
                    \EndIf
                \EndIf
            \EndFor
        \EndFor
    \EndFor
    \State $V_{\mathrm{ROA}}(l) \gets  \frac{1}{|\Omega|} \sum_{\omega=1}^{|\Omega|} F_{z_1}
    (t_0,X_{z_1,t_0}(\omega))$
    \State \Return $V_{\mathrm{ROA}}(l)$ 
\end{algorithmic}
\end{algorithm}

We propose Algorithm~1 that implements a dual backward recursion for achieving sequence evaluation according to ROA. Consider an investment sequence $l=\{z_1,z_2,\ldots,z_H\}$ consisting of $H$ portfolios. For each time step $t_n\in \mathcal{T}$, $|\Omega|$ MC paths are simulated to approximate the stochastic evolution of system states.
The algorithm first specifies terminal boundary conditions for each portfolio $z_h$. Owing to the constraints that (i) all $H$ portfolios must be invested sequentially within the planning horizon $\mathcal{T}$, and (ii) at most one portfolio can be invested at each time step, each portfolio is associated with a portfolio-specific latest admissible investment time $t_e$. Accordingly, prior to backward recursion, the stopping time of each portfolio is initialized as $\tau_{z_h}=t_e$ across all MC paths. The terminal option value $F_{z_h}(t_e,X_{z_h,t_e})$ is then computed as the immediate payoff from investing $z_h$ at $t_e$, which depends on the state variables $X_{z_h,t_e}$, intra- and inter-period costs $(c_{\text{intra}},c_{\text{inter}})$, and the spillover effect $\eta$.

The core of Algorithm~1 is the dual backward recursion, which jointly determines the optimal stopping times and option values for all portfolios in the sequence. The recursion proceeds backward along both the time dimension and the portfolio order. At each time step and for each portfolio eligible for investment, the immediate payoff is first evaluated for all MC paths. For each portfolio except the last one in the sequence, the expected discounted value of the subsequent portfolio and the continuation value are estimated via LSMC regression, which ensures almost sure convergence as the number of simulation paths and basis functions increases \parencite{Longstaff_2001_Valuing,Stentoft_2004_Convergence}. The investment decision is made by comparing the value of immediate exercise with the continuation value. If immediate investment is optimal, the stopping time and option value are updated accordingly, and the stopping times of subsequent portfolios are adjusted to preserve the prescribed investment order. Otherwise, the option value is discounted and carried backward in time. The option value of the entire sequence $l$ ($V_{\mathrm{ROA}}(l)$) is finally obtained as the expected option value of the first portfolio $z_1$ at the initial time $t_0$, averaged over all MC paths.  

\subsection{Deep reinforcement learning (DRL) sequence generator}
\label{sec:4.2}
\begin{figure}[htbp]
    \centering
    \includegraphics[width=\textwidth]{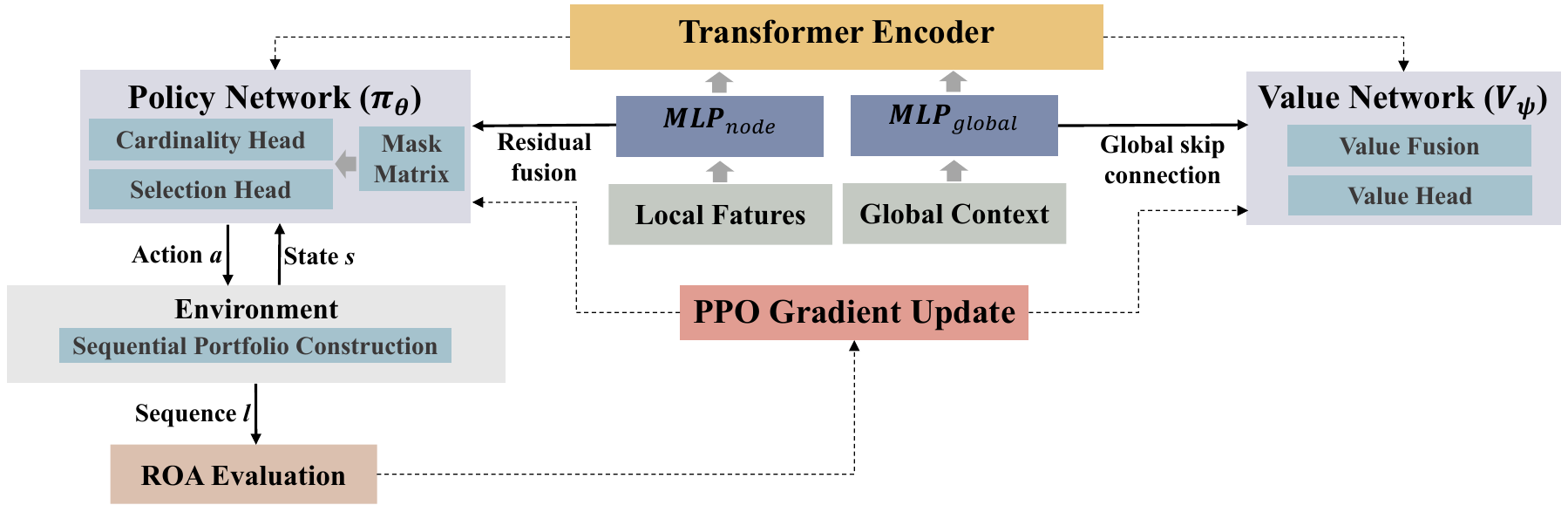}
    \caption{Framework of the TPPO}
    \label{fig:fig2}
\end{figure}

Enumerating and evaluating all feasible investment sequences using ROA quickly becomes computationally prohibitive in multi-region settings due to the combinatorial explosion of the sequence space. To overcome this challenge, we develop a DRL sequence generator that directly learns high option-value investment sequences.  The remainder of this section is organized as follows. Subsection~\ref{sec:4.2.1} formulates the sequence generation as a finite-horizon MDP. Subsection~\ref{sec:4.2.2} introduces the proposed Transformer-based Proximal Policy Optimization (TPPO) algorithm for sequence learning. The framework of the TPPO is illustrated in Figure~\ref{fig:fig2}.

\subsubsection{MDP formulation for sequence generation}
\label{sec:4.2.1}

We formulate the generation of investment sequences as a finite-horizon MDP defined over the discrete time steps $\mathcal{T}=\{t_0,t_1,\dots,t_T\}$. At each time step $t_n\in\mathcal{T}$, the agent may select at most one investment portfolio. Let $h$ denote the number of portfolios selected up to time $t_n$, and let $l_h=\{z_1,\dots,z_h\}$ denote the partially constructed sequence. Since at most one portfolio can be selected at each time step, the final sequence length satisfies $H\le T$.

The MDP is defined by $(\mathcal{S},\mathcal{A},P,R,\gamma)$ as follows.

\textbf{States ($\mathcal{S}$).}  
The state at time step $t_n$, denoted $s_n$, consists of:
(i) the investment status of each region in $\mathcal{N}$,
(ii) the partially constructed sequence $l_h=(z_1,\dots,z_h)$,
(iii) the current decision index $n$, equivalently the remaining time steps $T-n$.
The state summarizes all information required for subsequent portfolio selection and captures path dependency in sequential investment decisions.

\textbf{Actions ($\mathcal{A}$).}  
At time step $t_n$, the action selects the next portfolio $z_{h+1}\subseteq\mathcal{N}$ from the set of currently uninvested regions, subject to $|z_{h+1}|\le k$. 
For computational implementation, the action is represented as a binary vector $a_n\in\{0,1\}^{|\mathcal{N}|}$, where $a_{n,i}=1$ indicates that region $i$ is included in $z_{h+1}$. 
Feasibility is enforced by masking already-invested regions and restricting $\sum_{i\in\mathcal{N}} a_{n,i}\le k$.

\textbf{Transition dynamics ($P$).}  
State transitions are deterministic. 
Given state $s_n$ and action $a_n$ corresponding to portfolio $z_{h+1}$, the next state $s_{n+1}$ is obtained by appending $z_{h+1}$ to the sequence, i.e., $l_{h+1}=(z_1,\dots,z_h,z_{h+1})$, and updating the investment status of the selected regions. Demand uncertainty is not modeled within the MDP and is incorporated ex post through ROA evaluation.

\textbf{Rewards ($R$).}  
Let $l_h=\{z_1,\dots,z_h\}$ denote the partial sequence constructed up to time step $t_n$, with $l_0=\emptyset$. The sequence-level objective is the ROA-evaluated option value $V_{\mathrm{ROA}}(l_H)$ for the completed sequence $l_H\in\mathcal{L}$. To mitigate reward sparsity, learning is driven by marginal rewards derived from ROA evaluations:
\begin{equation}
r_n = V_{\mathrm{ROA}}(l_{h+1}) - V_{\mathrm{ROA}}(l_h),
\end{equation}
which represents the incremental contribution of appending portfolio $z_{h+1}$ to the sequence.

\textbf{Discount factor ($\gamma$).}  
The discount factor $\gamma\in(0,1]$ governs the temporal aggregation of step rewards over the finite horizon.

\subsubsection{Transformer-based Proximal Policy Optimization}
\label{sec:4.2.2}

We develop a Transformer-based Proximal Policy Optimization (TPPO) algorithm that integrates the relational reasoning of Transformers to capture inter-regional dependencies, employing structural augmentations to stabilize training.

\paragraph{\textbf{Hierarchical state embedding}}
The state $s_n$ is hierarchically decomposed into local node features $\mathbf{x}_{n,i} \in \mathbb{R}^4$ and global context $\mathbf{g}_n \in \mathbb{R}^3$, defined as $\mathbf{x}_{n,i} = [\mathbb{I}_{n,i}, \nu_n, \bar{t}_n, Q_{ii, t_0}]$ and $\mathbf{g}_n = [\bar{t}_n, \nu_n, \bar{Q}_{t_0}]$. Here, $\mathbb{I}_{n,i}$ indicates investment status, $\nu_n$ is the aggregate investment ratio, and $\bar{t}_n$ is the normalized time step. $Q_{ii, t_0}$ represents the baseline intra-region demand, while $\bar{Q}_{t_0}$ denotes the regional mean. \footnote{While dynamic inter-region (OD) demands evolve during the process, we explicitly utilize this static intra-region baseline as a proxy for each region's intrinsic market potential to maintain a compact state representation.} We employ a Transformer encoder to capture regional dependencies. However, the inherent permutation invariance of standard self-attention treats inputs as unordered sets, failing to distinguish the unique, time-invariant spatial identities of specific regions required for consistent tracking. To address this, we project local features via a Multi-Layer Perceptron (MLP) and inject a learnable region identity embedding $E_{id}(i)$, yielding the latent node token $\mathbf{o}_{n,i}^{(0)} = \text{MLP}_{node}(\mathbf{x}_{n,i}) + E_{id}(i)$. Regarding global aggregation, standard mean-pooling indiscriminately averages node features, often diluting sparse high-value signals. To avoid this, we prepend a classification token initialized from the global context: $\mathbf{o}_{cls, n}^{(0)} = \text{MLP}_{global}(\mathbf{g}_n) + \mathbf{w}_{cls}$, where $\mathbf{w}_{cls}$ is a learnable parameter. Unlike fixed pooling, this token utilizes self-attention to dynamically prioritize relevant regions.

\paragraph{\textbf{Transformer-based policy network}}
The policy network $\pi_\theta$ processes the token sequence $\mathcal{O}_n^{(0)}$ through $L$ self-attention layers, yielding representations $\mathbf{o}_{cls, n}^{(L)}$ and $\{\mathbf{o}_{n,i}^{(L)}\}_{i \in \mathcal{N}}$. We employ a dual-head decoding structure. The quantity head maps the global token to an investment size distribution: $\pi_k(k \mid s_n) = \text{Softmax}(\text{MLP}_{qh}(\mathbf{o}_{cls, n}^{(L)}))$. Simultaneously, the selection head uses residual feature fusion to retain direct access to raw state indicators. We concatenate re-encoded raw features with transformer outputs to compute selection logits $u_{n,i} = \mathbf{w}_{sq}^\top \mathbf{m}_{n,i}^{fuse}$, where $\mathbf{m}_{n,i}^{fuse} = \text{LayerNorm}(\text{ReLU}(\mathbf{W}_{fuse} [\mathbf{o}_{n,i}^{(L)}; \text{MLP}_{raw}(\mathbf{x}_{n,i})]))$. To enforce feasibility, logits of already invested regions are masked to $-\infty$ before Softmax, ensuring the sampled portfolio satisfies $\sum_{i \in z_{n+1}} \mathbb{I}_{n,i} = 0$.

\paragraph{\textbf{Symmetric critic with global skip connection}}
The critic $V_\psi$ shares a symmetric encoder architecture with the policy. However, since $V(s_n)$ is dominated by global variables (e.g., $\bar{t}_n$), passing these signals through deep attention layers is inefficient. We explicitly address this via a global skip connection in the value head: $V(s_n) = \text{MLP}_{vh}([\mathbf{o}_{cls, n}^{(L)}; \mathbf{g}_n])$. This decomposes value modeling, allowing the direct path to capture dominant linear trends (e.g., time decay) while the transformer captures complex residuals from regional interactions. 

We jointly update the policy parameters $\theta$ and critic parameters $\psi$ by maximizing the PPO objective with Generalized Advantage Estimation (GAE):
\begin{equation} 
    \max_{\theta, \psi} \mathbb{E}_n \left[ \min(r_n(\theta)\hat{A}_n, \text{clip}(r_n(\theta), 1-\epsilon, 1+\epsilon)\hat{A}_n) - c_1 (V_\psi(s_n) - V_n^{target})^2 + c_2 B(\pi_\theta(\cdot|s_n)) \right] 
\end{equation}
Here, $r_n(\theta) = \frac{\pi_\theta(a_n|s_n)}{\pi_{\theta_{old}}(a_n|s_n)}$ denotes the importance sampling probability ratio between the current and old policies, while $\hat{A}_n$ is the advantage estimate calculated via GAE. The clipping parameter $\epsilon$ restricts the policy update range to prevent destructive large steps. The objective simultaneously minimizes the value function's squared error against the bootstrapping target $V_n^{target}$ (scaled by coefficient $c_1$) and maximizes the policy entropy $B$ (scaled by $c_2$) to encourage exploration.

\section{Numerical Experiments}
\label{sec:5}

This section assesses the performance of the proposed DRL sequence generator, TPPO, through a series of computational experiments. Section 5.1 outlines the experimental setup based on two metropolitan multi-region datasets. Section 5.2 benchmarks TPPO against alternative DRL algorithms and other comparison methods. Section 5.3 presents sensitivity analyses and derives managerial insights. Section 5.4 provides an extended case study on the phased rollout of a mobility-on-demand (MoD) service in New York City under various planning scenarios. All experiments are implemented in Python 3.10 using PyTorch 2.0.1 and OpenAI Gym, with training conducted in a GPU-enabled environment (CUDA 11.7).

\subsection{Experimental Setup}
\label{sec:5.1}

 The experiments are conducted based on two realistic metropolitan cases: eight administrative regions from Shanghai and nine from Beijing, China. These regions are selected according to official geographic boundaries, allowing us to generate multiple multi-region configurations based on real-world spatial structures. Population density data are obtained from official statistics published by the National Bureau of Statistics of China \parencite{NBS2023} (see Table~\ref{tab:tableA1} in the Appendix). 
The planning horizon is set to $T = 5$ years, with annual decision epochs. The maximum number of regions that can be invested in per period, $k$, varies from 2 to 6 across experiments to reflect different budget constraints.

Initial regional demand at $t=0$ is calibrated based on structural characteristics: each region’s baseline demand is proportional to its geographic area and population density. The initial origin–destination (OD) matrix is constructed by allocating a fixed proportion of baseline demand to intra-regional flows, with the remainder uniformly distributed across inter-regional pairs. Demand dynamics incorporate heterogeneous growth, volatility, and the spillover effect determined by normalized density and area indices. The drift parameter $\mu_i$ increases with demand intensity but decreases with spatial coverage, reflecting stronger expansion potential in dense yet non-saturated regions. The volatility parameter $\sigma_i$ is higher for relatively small and dense regions, capturing greater sensitivity to localized fluctuations. The Poisson jump intensity $\lambda_i$ is region-specific and linearly mapped to the interval $[0.20,\,1.20]$ based on the interaction between normalized density and area indices.

During the investment process, we consider a positive spillover effect $\eta$ modeled by a Gamma distribution, $\eta \sim \Gamma(\xi_1, \xi_2)$, where $\xi_1 \sim U(0.1, 0.2)$ and $\xi_2 \sim U(0.4, 0.5)$ are randomly sampled shape and scale parameters to capture regional heterogeneity. Alternative specifications of $\eta$ are examined in the sensitivity analysis. For the ROA evaluation, we employ $J = 3$ basis functions, a discount rate of $\rho = 1\%$, and $|P| = 300$ Monte Carlo simulation paths. The intra-region cost $c_{\text{intra}}$ is set to 40\% of the average intra-regional demand at $t_0$, while the inter-region cost $c_{\text{inter}}$ equals 15\% of the average inter-regional demand at $t_0$.

\subsection{Algorithmic Performance}
\label{sec:5.2}

\subsubsection{Comparisons with alternative DRL algorithms}
\label{sec:5.2.1}

To evaluate the structural contribution of the Transformer architecture in TPPO, we compare our approach with alternative DRL algorithms, including the standard on-policy Proximal Policy Optimization (PPO) and the off-policy Soft Actor-Critic (SAC). SAC is adopted due to its maximum-entropy objective, which improves exploration by incorporating entropy regularization into the reward function. To isolate the architectural effect, we further implement a Transformer-augmented SAC variant (TSAC), which adopts the same hierarchical state embedding and encoder structure as TPPO while retaining SAC’s off-policy optimization scheme. All algorithms are trained under aligned hyperparameters to ensure a fair comparison. Experiments are conducted under two settings: a 7-region instance with $k=3$ (1,000 training episodes) and a 9-region instance with $k=4$ (2,000 training episodes).

\begin{figure}[htbp]
    \centering
    \includegraphics[width=0.6\textwidth]{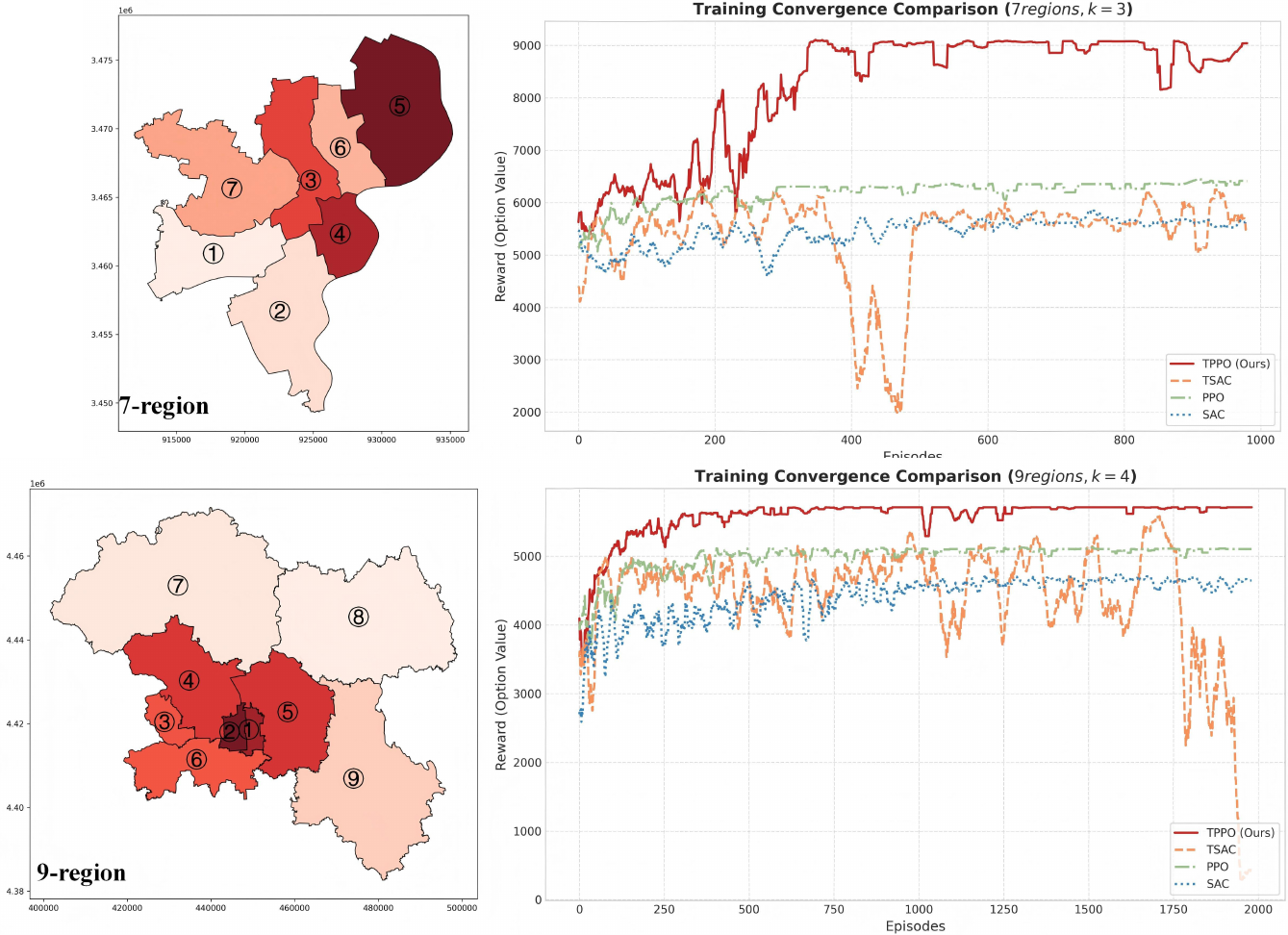}
    \caption{Training curves of TPPO vs. alternative DRL approaches}
    \label{fig:fig3}
\end{figure}

As illustrated in Figure~\ref{fig:fig3}, TPPO consistently achieves faster convergence and higher asymptotic rewards across both scenarios. This indicates that TPPO more effectively captures inter-regional dependencies and learns investment sequences with greater option value. Notably, the on-policy methods (TPPO and PPO) exhibit higher stability and superior final performance than the off-policy approaches (SAC and TSAC), likely due to the sequential and combinatorial structure of the decision process, where on-policy updates better preserve trajectory consistency. Moreover, the absence of performance gains in TSAC suggests that the effectiveness of the Transformer encoder depends on its interaction with the optimization paradigm: without PPO’s clipped objective to stabilize policy updates, the complex relational representations may increase gradient variance and limit off-policy convergence.

\subsubsection{Comparisons with the enumeration approach}
\label{sec:5.2.2} 
We first apply the ROA framework to six small-scale instances constructed from subsets of districts in Shanghai and Beijing, with the number of regions ranging from 4 to 6 and investment capacity $k \in \{2,4\}$. Table~\ref{tab:table2} reports the results. As the problem size increases, the number of feasible investment sequences grows rapidly—from 66--75 in 4-region instances to 450--540 in 5-region cases, and further to 2,970--3,950 in 6-region settings. This combinatorial expansion leads to a sharp increase in computational effort: average runtime rises from under 10 seconds to over 100 seconds and exceeds 1200 seconds as the number of regions increases from 4 to 6. Additionally, a larger investment capacity $k$ further enlarges the solution space and exacerbates computational complexity. The option value distributions show substantial dispersion across feasible sequences, indicating that investment ordering critically affects performance and making high-value solutions difficult to identify through enumeration alone as problem size grows. Notably, the highest-value sequences remain structurally stable across different $k$ settings within the same scenario. In several cases, increasing $k$ neither changes the optimal ordering nor leads to full utilization of the maximum allowable simultaneous investments. Instead, the best-performing strategies invest sequentially, typically one region at a time and at most two concurrently, highlighting the importance of timing and staged expansion over aggressive early deployment.

\begin{table}[htbp]
\centering
\resizebox{\textwidth}{!}{%
\begin{tabular}{cccccccc} 
\toprule 
Scenario & Region volume & k & Number of sequences & Sequence with highest option value ($s_h$) & Option value of $s_h$ & Runtime(s) & Sequence option value distribution \\ \midrule

% --- 第一组数据 ---
% 关键修改 2: 文件名去掉了空格，与截图保持一致 (FigureT3.1.pdf)
\multirow{2}{*}{\includegraphics[width=2cm, valign=c]{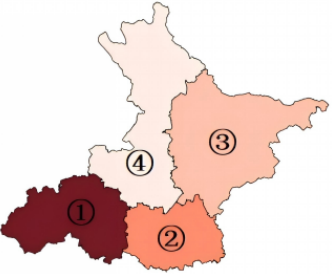}} & 
\multirow{2}{*}{\begin{tabular}[c]{@{}c@{}}4\\ (Shanghai)\end{tabular}} & 
2 & 66 & \{[r4], [r2], [r1], [r3]\} & 400.34 & 9.13 & \includegraphics[width=3cm, valign=c]{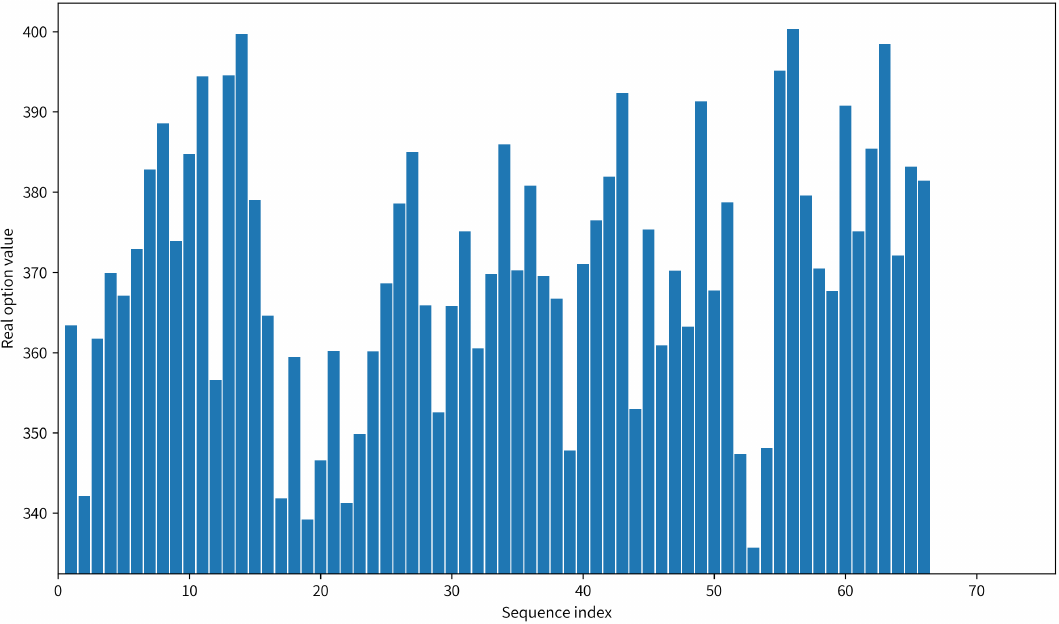} \\
 &  & 4 & 75 & \{[r4], [r2], [r1], [r3]\} & 400.34 & 9.96 & \includegraphics[width=3cm, valign=c]{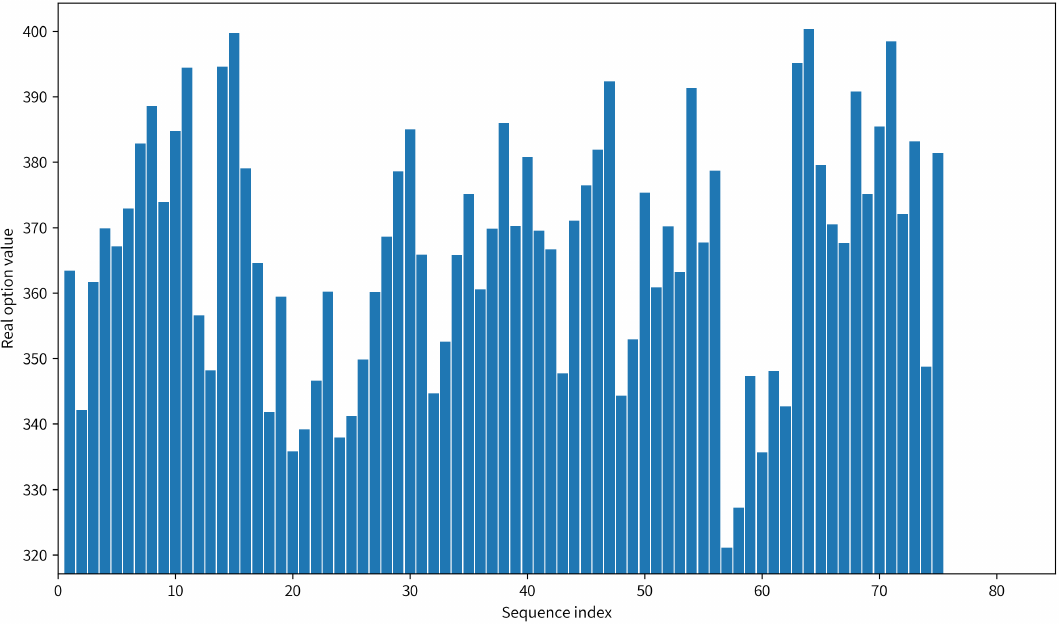} \\ \midrule

% --- 第二组数据 ---
\multirow{2}{*}{\includegraphics[width=2cm, valign=c]{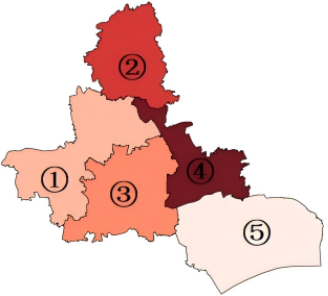}} & 
\multirow{2}{*}{\begin{tabular}[c]{@{}c@{}}5\\ (Shanghai)\end{tabular}} & 
2 & 450 & \{[r3], [r5], [r2], [r4], [r1]\} & 415.52 & 101.61 & \includegraphics[width=3cm, valign=c]{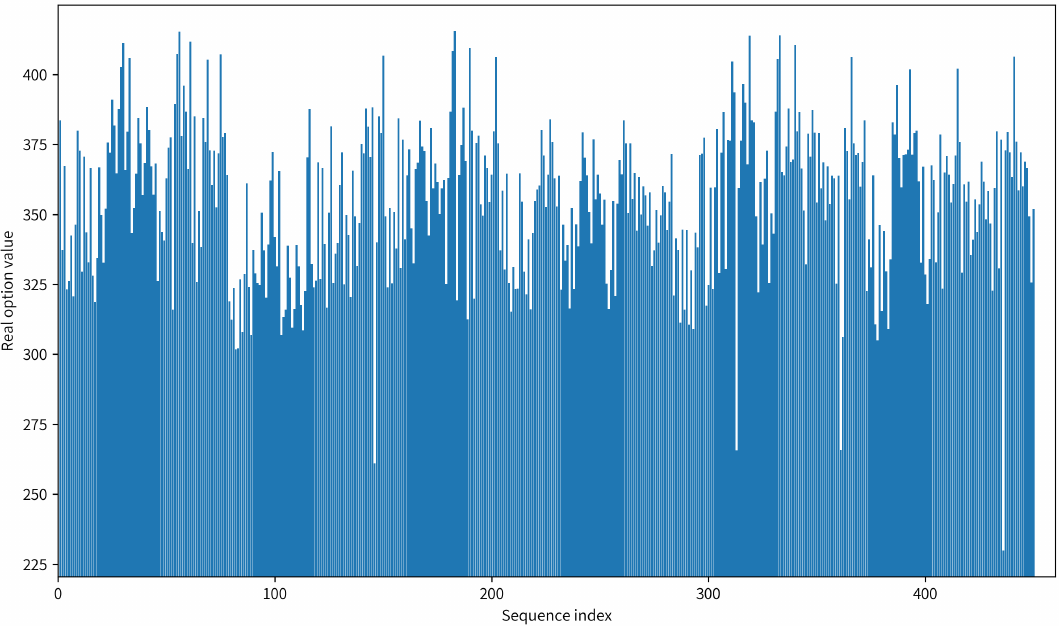} \\
 &  & 4 & 540 & \{[r5], [r3], [r2], [r4], [r1]\} & 386.62 & 115.43 & \includegraphics[width=3cm, valign=c]{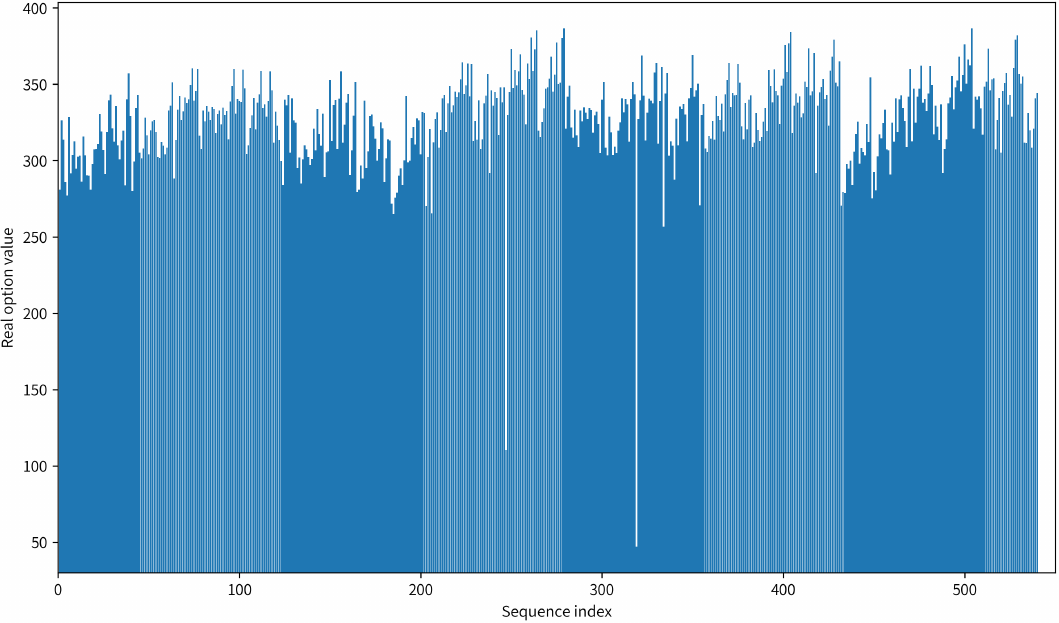} \\ \midrule

% --- 第三组数据 ---
\multirow{2}{*}{\includegraphics[width=2cm, valign=c]{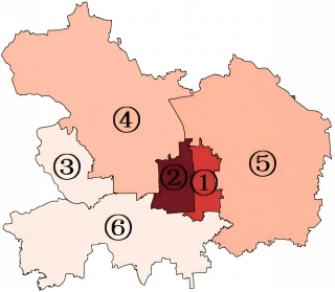}} & 
\multirow{2}{*}{\begin{tabular}[c]{@{}c@{}}6\\ (Beijing)\end{tabular}} & 
2 & 2970 & \{[r1], [r4], [r5], [r2], [r3,r6]\} & 4334.64 & 853.33 & \includegraphics[width=3cm, valign=c]{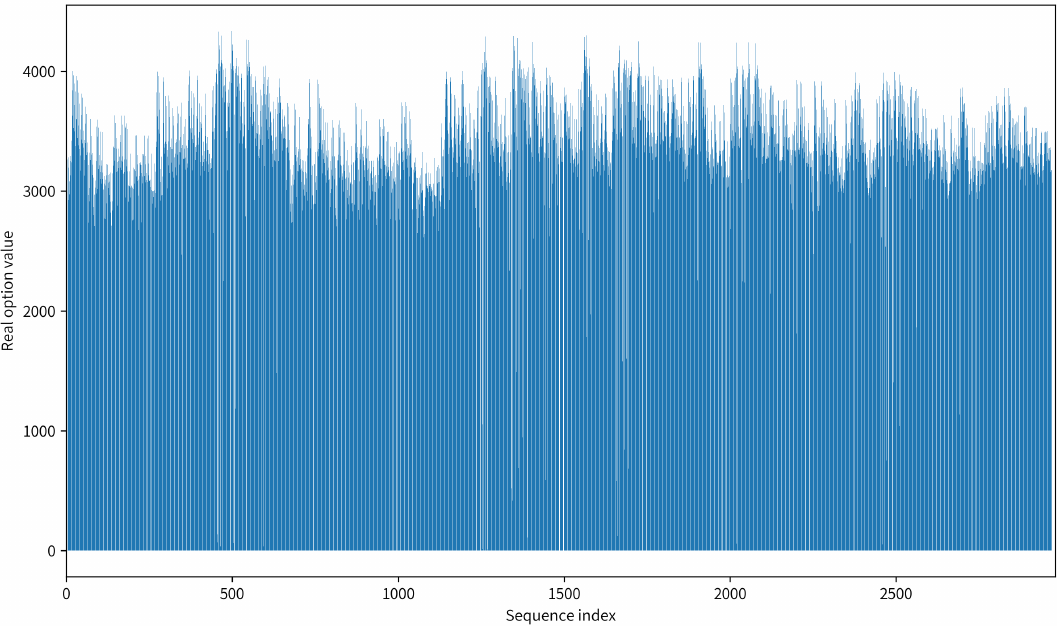} \\
 &  & 4 & 3950 & \{[r4], [r1], [r5], [r2], [r3, r6]\} & 4454.49 & 1232.31 & \includegraphics[width=3cm, valign=c]{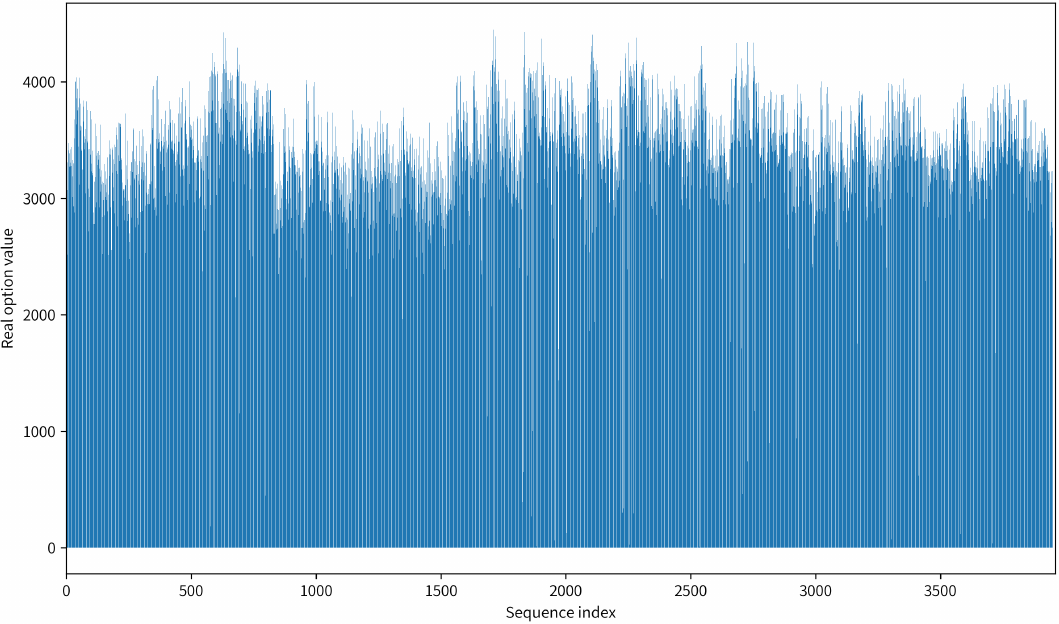} \\ \bottomrule
\end{tabular}%
}
\caption{Comparison with the enumeration approach (4-6 region map)}
\label{tab:table2}
\end{table}

We next evaluate 6-region and 7-region instances from Shanghai with investment capacity $k \in \{2,3\}$, where TPPO is trained for 500 episodes and benchmarked against exhaustive enumeration (Table~\ref {tab:table3}). As the problem size increases, the number of feasible sequences grows rapidly, from 2970--3830 in 6-region cases to 15,120 and 25,410 in 7-region instances for $k=2$ and $k=3$, respectively, leading to a sharp increase in enumeration time from approximately 1000 seconds to over 11000 seconds. Despite this combinatorial growth, TPPO achieves near-optimal performance with substantially lower computational cost, yielding an average optimality gap of only 1.31\% while reducing runtime significantly. In particular, TPPO exactly matches the optimal solution in Ins.~6-2 with about one-third of the runtime, and attains a gap of only 0.06\% in Ins.~7-2 using less than 5\% of the enumeration time. Overall, TPPO maintains stable accuracy and runtime across all instances, demonstrating strong scalability for large-scale investment sequencing problems.

\begin{table}[htbp]
\centering
% 1. 设置更小的字体 (可选 \small, \footnotesize, \scriptsize)
\footnotesize 
% 2. 减小列与列之间的间距 (默认是 6pt，改为 3pt)
\setlength{\tabcolsep}{3pt} 

\begin{tabular}{@{} l c c c c c c @{}}
\toprule
 &  & \multicolumn{3}{c}{TPPO} & \multicolumn{2}{c}{Enumeration} \\ \cmidrule(lr){3-5} \cmidrule(l){6-7} 
\multirow{-2}{*}{Instances} & \multirow{-2}{*}{\begin{tabular}[c]{@{}c@{}}Num of\\ seqs\end{tabular}} & Option value & Gap & Runtime(s) & Option value & Runtime(s) \\ \midrule
\textit{Ins.6-2} & 2970 & 4626.01 & 0.00\% & 309.32 & 4626.01 & 997.60 \\
\textit{Ins.6-3} & 3830 & 4521.48 & 2.26\% & 278.01 & 4626.01 & 1232.83 \\
\textit{Ins.7-2} & 15120 & 7558.69 & 0.06\% & 324.47 & 7563.24 & 6892.42 \\
\textit{Ins.7-3} & 25410 & 7274.70 & 2.93\% & 282.83 & 7494.53 & 11164.09 \\ 
\multicolumn{2}{c}{Average} & 5995.22 & 1.31\% & 298.66 & 6077.45 & 5071.74 \\ \midrule
% 下半部分
\multicolumn{2}{c}{Instances} & \multicolumn{3}{c}{TPPO} & \multicolumn{2}{c}{Enumeration} \\ \midrule
% 这里使用了 \scriptsize 专门针对这些很长的序列文字，防止表格被撑得太宽
\multicolumn{2}{c}{\textit{Ins.6-2}} & \multicolumn{3}{l}{\scriptsize \{[r4], [r1], [r5], [r2], [r3, r6]\}} & \multicolumn{2}{l}{\scriptsize \{[r4], [r1], [r5], [r2], [r3, r6]\}} \\
\multicolumn{2}{c}{\textit{Ins.6-3}} & \multicolumn{3}{l}{\scriptsize \{[r4], [r1], [r2, r5], [r6], [r3]\}} & \multicolumn{2}{l}{\scriptsize \{[r4], [r1], [r5], [r2], [r3, r6]\}} \\
\multicolumn{2}{c}{\textit{Ins.7-2}} & \multicolumn{3}{l}{\scriptsize \{[r1], [r4], [r2], [r3, r6], [r5, r7]\}} & \multicolumn{2}{l}{\scriptsize \{[r1], [r4], [r6], [r2, r7], [r3, r5]\}} \\
\multicolumn{2}{c}{\textit{Ins.7-3}} & \multicolumn{3}{l}{\scriptsize \{[r1], [r4], [r2], [r3, r6], [r5, r7]\}} & \multicolumn{2}{l}{\scriptsize \{[r1], [r4], [r6], [r2], [r3, r5, r7]\}} \\ \bottomrule
\end{tabular}

\caption{Option value results for TPPO vs enumeration approach Pretrained (500 episodes)}
\label{tab:table3}
\end{table}

\subsubsection{Comparisons with alternative heuristics}
\label{sec:5.2.3} 

To evaluate the performance and scalability of the TPPO framework, we conduct benchmark tests across nine scenarios involving 7, 8, and 9 regions, utilizing geographical data from Shanghai (7- and 8-region) and Beijing (9-region). The TPPO model is trained for 1000 episodes per scenario, and its performance is compared against two baseline myopic heuristics: Myopia-H and Myopia-L, which prioritize investments based on high and low initial demand, respectively. For each instance ($k = 3, 4, 5$), we report the best-found sequence from 20 runs, with option values calculated via ROA. 

\begin{table}[htbp]
\centering
% 1. 使用 \small 或 \footnotesize 让字体变小
\small
% 2. 稍微减小列间距 (默认是6pt，改为4pt)
\setlength{\tabcolsep}{4pt}

\begin{tabular}{@{} l c c c c c @{}}
\toprule
% 表头优化：Instances 跨两行，Option Value 跨后五列
\multirow{2}{*}{Instances} & \multicolumn{5}{c}{Option value} \\ \cmidrule(l){2-6} 
 & TPPO & Myopia-L & Imv. & Myopia-H & Imv. \\ \midrule
\textit{Ins.7-3} & 7568.45 & 5813.13 & 30.20\% & 4862.98 & 55.63\% \\
\textit{Ins.7-4} & 7401.21 & 5802.00 & 27.56\% & 5161.72 & 43.39\% \\
\textit{Ins.7-5} & 7213.03 & 5915.72 & 21.93\% & 4142.68 & 74.12\% \\
\textit{Ins.8-3} & 7211.72 & 6874.55 & 4.90\% & 4068.75 & 77.25\% \\
\textit{Ins.8-4} & 7452.72 & 7307.80 & 1.98\% & 5444.05 & 36.90\% \\
\textit{Ins.8-5} & 7257.88 & 7052.03 & 2.92\% & 3158.18 & 129.81\% \\
\textit{Ins.9-3} & 5771.83 & 4676.09 & 23.43\% & 4327.60 & 33.37\% \\
\textit{Ins.9-4} & 5173.29 & 4620.34 & 11.97\% & 4586.63 & 12.79\% \\
\textit{Ins.9-5} & 5824.97 & 5383.31 & 8.20\% & 4406.21 & 32.20\% \\
Average & 6763.90 & 5938.33 & 13.90\% & 4462.09 & 51.59\% \\ \bottomrule
\end{tabular}
% 修正了 caption 中的拼写错误 (benchmarks)
\caption{Option value results for TPPO vs alternative benchmarks}
\label{tab:table4}
\end{table}

\begin{figure}[htbp]
    \centering
    \includegraphics[width=0.6\textwidth]{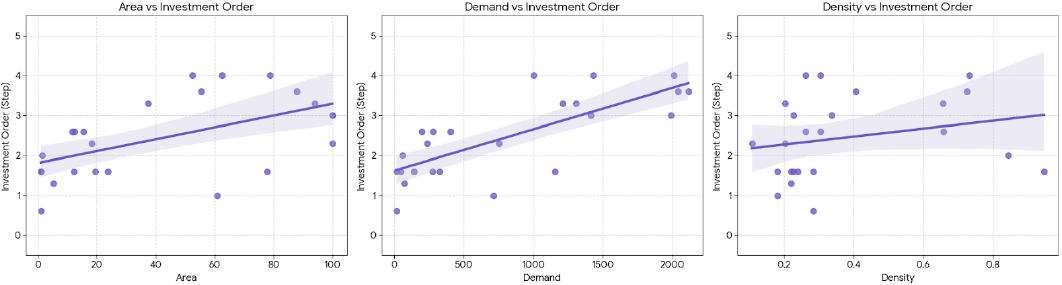}
    \caption{Correlation analysis between investment order and regional attributes}
    \label{fig:fig4}
\end{figure}

As detailed in Table~\ref{tab:table4}, TPPO consistently outperforms both heuristics across all test instances. On average, TPPO delivers option values that are 13.90\% higher than Myopia-L and 51.59\% higher than Myopia-H. While Myopia-L demonstrates relative competitiveness in certain 8-region instances, maintaining a performance gap of less than 3\%, its fixed-ranking approach fails to account for long-term demand uncertainty. Across all scenarios, the investment strategy of TPPO follows a "bottom-up" approach: the early investment phase focuses on regions with a small average area and low initial demand to achieve "quick wins". In contrast, regions with the highest initial demand are consistently deferred to the final stages of the sequence. Figure~\ref{fig:fig4} confirms that the decision to invest early is driven primarily by a region's area and initial demand rather than high density alone. This is evidenced by a strong correlation ($0.65$) between demand and investment timing, compared to a negligible correlation ($0.05$) for density. 

\subsection{Practical insights from sensitivity analysis}
\label{sec:5.3}
This section uses the 9-region Beijing scenario to conduct sensitivity analyses on two key factors: the investment capacity constraint $k$ and the spillover effect. We examine how variations in these factors influence the performance of TPPO and extract managerial insights for sequential service region design.

\subsubsection{Analysis of k-region constraint}
\label{sec:5.3.1}
We evaluate nine demand regimes by varying the drift $\mu$ and volatility $\sigma$ across low, medium, and high levels (e.g., Ins.9-LL denotes low $\mu$ and low $\sigma$, whereas Ins.9-MH represents medium $\mu$ and high $\sigma$). Table~\ref{tab:table5} reports the option values of the best sequences identified by TPPO (within 500 training episodes) and the two myopic heuristics after 20 runs under different investment capacities $k \in \{2,\dots,6\}$.

\begin{table}[htbp]
\centering
% 强制压缩到页面宽度
\resizebox{\textwidth}{!}{%
% 极度压缩列间距
\setlength{\tabcolsep}{2pt}

\begin{tabular}{@{} l ccc ccc ccc ccc ccc @{}}
\toprule
 & \multicolumn{15}{c}{Option value} \\ 
\cmidrule(l){2-16} 
 & \multicolumn{3}{c}{k=2} & \multicolumn{3}{c}{k=3} & \multicolumn{3}{c}{k=4} & \multicolumn{3}{c}{k=5} & \multicolumn{3}{c}{k=6} \\ 
\cmidrule(lr){2-4} \cmidrule(lr){5-7} \cmidrule(lr){8-10} \cmidrule(lr){11-13} \cmidrule(l){14-16}

Instances & TPPO & Myopia-H & Myopia-L & TPPO & Myopia-H & Myopia-L & TPPO & Myopia-H & Myopia-L & TPPO & Myopia-H & Myopia-L & TPPO & Myopia-H & Myopia-L \\ \midrule
\textit{Ins.9-LL} & 5719.03 & 4525.69 & 5158.77 & 5643.06 & 4523.64 & 3645.08 & 5779.15 & 3614.57 & 4699.59 & 5712.58 & 3878.46 & 3594.28 & 5143.12 & 4446.70 & 4889.88 \\
\textit{Ins.9-LM} & 5739.52 & 3945.11 & 879.29 & 5603.96 & 4266.21 & 4625.47 & 5717.22 & 4114.15 & 4835.90 & 6246.70 & 4863.29 & 3067.16 & 5274.35 & 4309.31 & 4244.96 \\
\textit{Ins.9-LH} & 5674.93 & 3989.11 & 3113.27 & 584.04 & 3927.68 & 2907.94 & 5917.70 & 4415.96 & 4743.75 & 5997.32 & 2759.30 & 4997.32 & 6201.23 & 3780.50 & 4084.02 \\
\textit{Ins.9-ML} & 6051.71 & 4607.05 & 5485.16 & 6267.57 & 4718.12 & 4917.96 & 6260.23 & 4022.52 & 3482.12 & 6464.91 & 4542.25 & 3745.13 & 5920.48 & 4875.61 & 3760.32 \\
\textit{Ins.9-MM} & 6005.46 & 4461.48 & 4578.98 & 6310.99 & 4305.60 & 5511.36 & 6414.36 & 4022.98 & 5100.30 & 5948.63 & 4910.46 & 3169.76 & 6540.00 & 4244.00 & 5047.47 \\
\textit{Ins.9-MH} & 6004.26 & 4322.12 & 4445.05 & 6239.24 & 4498.65 & 3515.88 & 6009.05 & 3739.40 & 5073.17 & 6340.65 & 5510.48 & 3175.64 & 6214.35 & 5074.78 & 2381.78 \\
\textit{Ins.9-HL} & 6436.52 & 4194.78 & 6050.51 & 6707.91 & 4663.92 & 4678.22 & 6464.02 & 4741.80 & 4978.06 & 6464.98 & 4915.83 & 3336.36 & 6678.81 & 4597.80 & 3502.40 \\
\textit{Ins.9-HM} & 6223.00 & 4718.10 & 5411.90 & 6695.01 & 4718.10 & 5325.17 & 6441.54 & 3872.72 & 4739.34 & 6489.48 & 4376.93 & 3292.03 & 6435.64 & 5271.35 & 5432.49 \\
\textit{Ins.9-HH} & 6308.19 & 4610.38 & 4691.99 & 6308.19 & 4826.74 & 4163.28 & 6401.05 & 4433.02 & 4422.72 & 6694.51 & 4799.81 & 4060.24 & 6184.17 & 4731.71 & 3806.26 \\ 
Average & 6018.07 & 4374.87 & 4962.70 & 6180.00 & 4494.30 & 4455.66 & 6156.04 & 4108.57 & 4675.00 & 6295.30 & 4506.31 & 3604.21 & 6065.79 & 4592.42 & 4127.73 \\ 
Imv. & - & 37.56\% & 21.27\% & - & 37.51\% & 38.70\% & - & 49.83\% & 31.68\% & - & 39.70\% & 74.67\% & - & 32.08\% & 46.95\% \\ 
\bottomrule
\end{tabular}
}
\caption{Sensitivity analysis of k-region constraint)}
\label{tab:table5}
\end{table}

Across all demand regimes and $k$ levels, TPPO consistently outperforms the two myopic heuristics. On average, TPPO achieves improvements ranging from 32\% to nearly 50\% over Myopia-H and from 21\% to over 70\% over Myopia-L, with particularly pronounced gains under higher growth and volatility scenarios. The results show that relaxing the $k$-region constraint generally increases option value by enhancing strategic flexibility, although the effect is non-monotonic. Taking TPPO as an example, the average option value increases from 6,018.07 at $k=2$ to 6295.30 at $k=5$, before slightly declining at $k=6$. This pattern indicates diminishing returns to investment concurrency: increasing $k$ enhances strategic flexibility and allows more diversified portfolio deployment, but excessive simultaneous expansion reduces intertemporal flexibility and weakens the real-option value of the overall sequence. Importantly, the impact of the $k$-region constraint depends on the underlying demand dynamics. In relatively stable environments (e.g., Ins.9-LL), increasing $k$ provides only marginal benefits and may even reduce option value at larger $k$, as the value of flexibility under uncertainty is limited. In contrast, under higher growth or volatility (e.g., Ins.9-MH and Ins.9-HL), moderate increases in $k$ generate more substantial gains. Consistent with earlier observations, performance across most scenarios peaks at intermediate levels ($k=4$ or 5), indicating diminishing returns to aggressive expansion. From a managerial perspective, investment concurrency should align with market conditions: conservative pacing (small $k$) suffices in stable environments, whereas moderate parallel deployment ($k=4$ or $5$) is preferable in more dynamic markets. Fully aggressive expansion ($k=6$) is rarely optimal.

\begin{figure}[htbp]
    \centering
    \includegraphics[width=0.8\textwidth]{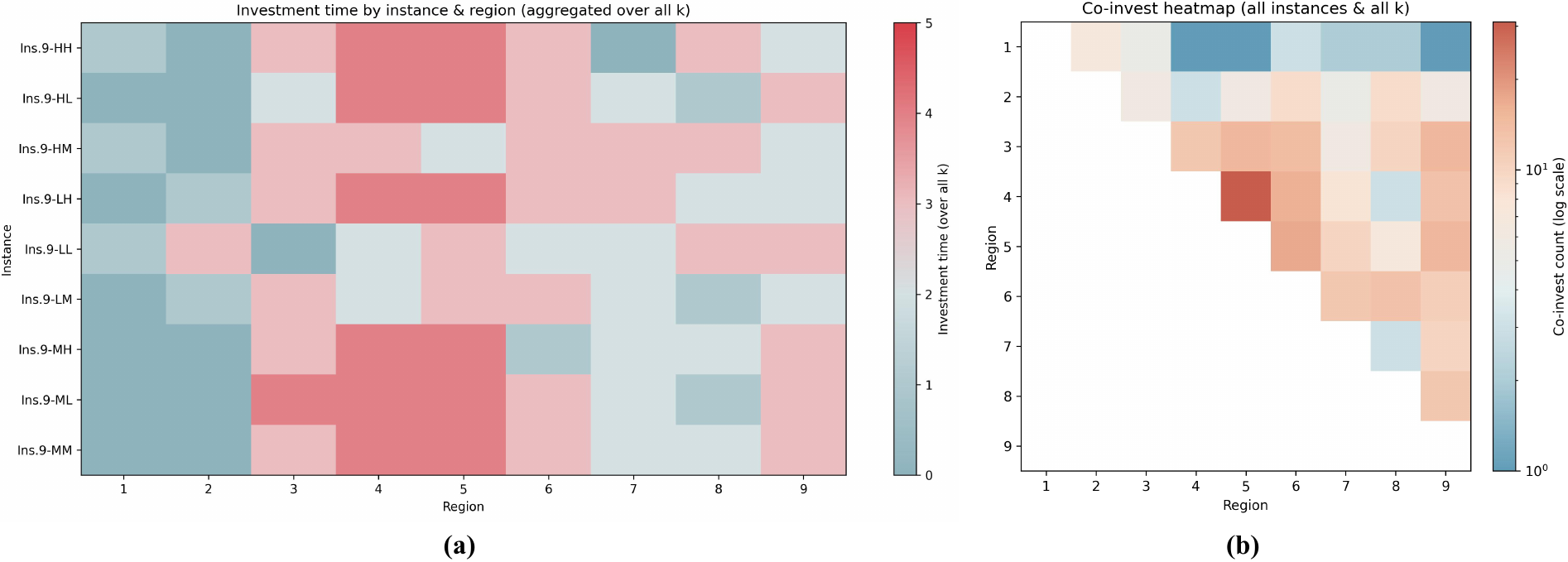}
    \caption{Heatmap analysis of k-region constraint}
    \label{fig:fig5}
\end{figure}

The investment time heatmap (Figure~\ref{fig:fig5}(a)) reveals a counterintuitive but economically consistent phenomenon. Although Regions 4, 5, and 9 exhibit the highest baseline demand, they are systematically invested later across most demand patterns and $k$ settings. In contrast, smaller or lower base-demand regions tend to be deployed earlier. This sequencing suggests that high-demand regions retain greater option value when deferred, as postponement preserves flexibility under uncertainty. Early-stage investments are therefore directed toward regions with a lower opportunity cost of waiting. The co-investment heatmap (Figure~\ref{fig:fig5}(b)) further shows that investment concurrency is selective rather than uniform. Regions 4 and 5 exhibit the highest co-invest frequency, with additional clustering observed for pairs such as 3\&4 and 5\&9. These regions share relatively high baseline demand, determined by the interaction of spatial area and population density. However, high baseline demand alone does not imply universal co-investment. For example, although Region 4 has a large baseline demand, it's not uniformly co-invested with all regions. Moreover, Region 1, despite having high population density, is rarely co-invested with Regions 4 or 5 due to its comparatively small total demand generated by a limited spatial area. Together, these patterns indicate that investment timing and concurrency are governed by marginal gains from joint deployment relative to the value of staging flexibility.

\subsubsection{Analysis of spillover effect}
\label{sec:5.3.2}
In the previous experiments, the stochastic spillover effect $\eta$ is modeled as Gamma-distributed, ensuring a positive spillover effect on regional demand. To examine the robustness of our results under alternative spillover structures, we consider three additional distributions: Lognormal, Normal, and Laplace, to represent different shapes and potential signs of spillover effects. While the Lognormal distribution maintains strictly positive spillovers, the Normal and Laplace distributions allow for both positive and negative realizations, thereby capturing scenarios in which regional interactions may either amplify or dampen demand. In addition, we distinguish between stationary spillover effects, which remain constant regardless of network expansion, and nonstationary spillover effects that intensify as additional regions are developed. To further assess sensitivity, we vary the spillover strength parameter across three levels (0.8, 1.0, and 1.2), representing weaker, baseline, and stronger spillovers, respectively. Experiments are conducted on the 7-region (Shanghai) scenario with $k=3$, comparing TPPO (trained for 500 episodes) against Myopia-H and Myopia-L. Table~\ref{tab:table6} reports the average option values of the best-found sequences over 20 runs. 

\begin{table}[htbp]
\centering
% 方法 1：调整缩放比例
% 将 \textwidth 改为 0.75\textwidth (即只占页面宽度的75%)
% 这样表格整体（包括字体）都会按比例缩小
\resizebox{0.75\textwidth}{!}{%
    \setlength{\tabcolsep}{3pt} % 稍微减小列间距
    \renewcommand{\arraystretch}{0.85} % 压缩行高 (默认是1.0)
    
    \begin{tabular}{@{} l l ccc | ccc @{}}
    \toprule
     &  & \multicolumn{3}{c}{\textbf{Stationary = False}} & \multicolumn{3}{c}{\textbf{Stationary = True}} \\ 
    \cmidrule(lr){3-5} \cmidrule(l){6-8} 
    Distribution & Strength & TPPO & Myopia-H & Myopia-L & TPPO & Myopia-H & Myopia-L \\ \midrule

    % === Gamma ===
    \multirow{3}{*}{Gamma} 
     & 0.8 & 11059.43 & 8811.84 & 13837.79 & 5319.36 & 4437.03 & 4741.43 \\
     & 1.0 & 18857.02 & 15381.02 & 29330.18 & 7114.86 & 5621.10 & 6223.50 \\
     & 1.2 & 34817.72 & 31848.86 & 32686.78 & 9360.42 & 7209.27 & 8314.89 \\ \cmidrule(l){2-8} 

    % === Lognormal ===
    \multirow{3}{*}{Lognormal} 
     & 0.8 & 14171.83 & 11892.32 & 12931.20 & 5404.62 & 4444.51 & 4748.73 \\
     & 1.0 & 49639.97 & 45135.96 & 49848.64 & 7516.41 & 5623.14 & 6386.78 \\
     & 1.2 & 248444.72 & 150896.86 & 246404.55 & 11340.55 & 7832.67 & 9114.28 \\ \cmidrule(l){2-8} 

    % === Normal ===
    \multirow{3}{*}{Normal} 
     & 0.8 & 14439.60 & 8761.16 & 10561.00 & 5307.91 & 4336.38 & 5150.45 \\
     & 1.0 & 16785.19 & 14005.53 & 17591.88 & 7029.97 & 5506.90 & 6802.33 \\
     & 1.2 & 27812.52 & 22788.28 & 28685.70 & 9287.74 & 6990.23 & 9073.82 \\ \cmidrule(l){2-8} 

    % === Laplace ===
    \multirow{3}{*}{Laplace} 
     & 0.8 & 18725.34 & 8558.03 & 10649.63 & 6260.76 & 4414.07 & 4385.35 \\
     & 1.0 & 37308.98 & 13575.20 & 18628.60 & 8514.43 & 5528.23 & 5580.06 \\
     & 1.2 & 81170.18 & 22213.46 & 35339.69 & 11694.00 & 7023.66 & 7154.51 \\ \midrule

    % === Summary ===
    \multicolumn{2}{c}{Average} & 47769.37 & 29489.04 & 42207.97 & 7845.92 & 5747.27 & 6473.01 \\
    \multicolumn{2}{c}{Imv.} & - & 61.99\% & 13.18\% & - & 36.52\% & 21.21\% \\ \bottomrule
    \end{tabular}%
}
\caption{Sensitivity analysis of spillover effect}
\label{tab:table6}
\end{table}

Across all configurations, TPPO demonstrates strong robustness and consistently outperforms the myopic benchmark policies. Under nonstationary spillovers, where the spillover effect intensifies with network expansion, TPPO achieves the highest average option value (47769.37), exceeding Myopia-H (29489.04) by 61.99\% and Myopia-L (42207.97) by 13.18\%. The advantage of TPPO becomes particularly pronounced under stronger spillover strength (1.2), where option values increase dramatically. This suggests that when demand interactions compound as the network grows, DRL-based sequencing is critical for capturing cumulative network benefits. Under stationary spillovers, overall option values are substantially lower (TPPO average 7845.92). Nevertheless, TPPO remains dominant, outperforming Myopia-H and Myopia-L by 36.52\% and 21.21\%, respectively. As spillover strength increases from 0.8 to 1.2, option values rise monotonically for all policies, but the relative gain of TPPO remains stable or widens slightly. The superiority of TPPO persists across all four distributions. In particular, even when spillovers are allowed to be negative (Normal and Laplace), TPPO maintains the largest option value in most strength settings. Regarding spillover strength, increasing the parameter from 0.8 to 1.2 substantially amplifies option values for all policies, especially under strictly positive distributions (Gamma and Lognormal).  Notably, the performance gap between TPPO and the myopic heuristics generally widens as spillover strength increases, suggesting that DRL-based sequencing becomes more valuable when the spillover effect intensifies.

\subsection{Extended Case study: MoD service area expansion in NYC}
\label{sec:5.4}
\begin{figure}[htbp]
    \centering
    \includegraphics[width=0.6\textwidth]{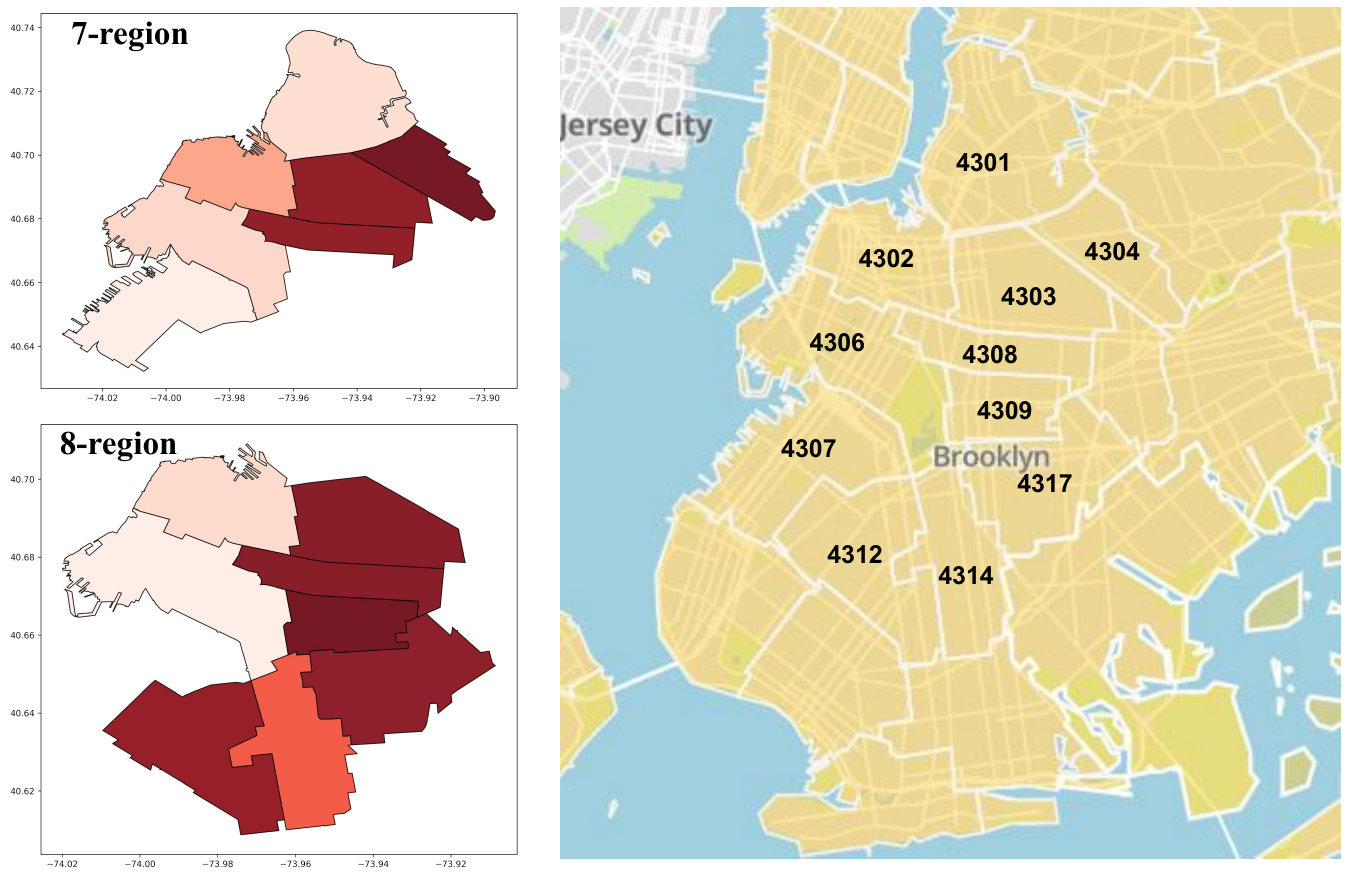}
    \caption{Multi-region scenarios in Brooklyn (KingsCounty), NYC}
    \label{fig:fig6}
\end{figure}

We extend the analysis to a MoD service area expansion problem in New York City (NYC). We construct 7- and 8-region scenarios in Brooklyn (Kings County), see Figure~\ref{fig:fig6}, where each region corresponds to a geographically contiguous Public Use Microdata Area (PUMA). The selected PUMAs are extracted from official 2020 Census GeoJSON files. Spatial attributes—including area (in km$^{2}$) and centroids—are computed after projection to a metric coordinate system. Population density for each PUMA is obtained from ACS (2019) data and normalized to ensure comparability across zones. Travel times between regions are calculated using centroid-to-centroid haversine distances with an assumed operating speed of $v=19.31 km/h$ and adjusted by peak-hour multipliers to reflect congestion conditions.

Potential MoD demand is simulated as follows: total daily commuting flows within the selected PUMAs are first aggregated, and peak-hour demand is approximated as 23.38\% of daily travel, corresponding to the average share of morning peak trips (7–9 AM) reported in ACS data for NYC. To reflect realistic market penetration, 60\% of peak-hour demand is treated as the upper bound of potential OD demand for MoD services. This value serves as the initial latent demand level at $t=0$. Regional baseline demand is proportional to the product of spatial area and population density. A fixed proportion (e.g., 0.3) of baseline demand remains intra-zonal, while the remaining demand is evenly distributed across inter-zonal OD pairs. Demand evolution follows a geometric Brownian motion with jump diffusion. The regional drift $\mu_i$ is linearly mapped to $[0.005,\,0.040]$ based on the interaction between normalized density and spatial capacity, while volatility $\sigma_i$ is mapped to $[0.18,\,0.55]$, assigning higher uncertainty to relatively small and dense zones. Other settings, including the spillover effect $\eta$, intra-region cost $c_{\text{intra}}$ and inter-region cost $c_{\text{inter}}$, follow the same settings as in Section~5.1. 
Realized ridership $\tilde{Q}_{ijt}$ is endogenously determined through a congestion-sensitive demand response:
\[
\tilde{Q}_{ijt} 
= Q_{ijt}
\exp\!\left(
-\delta 
\left[
p_{ij} 
+ \text{VOT} \left( TT_{ij} + T^{\text{wait}}_t \right)
\right]
\right),
\]
where $p_{ij}=2.42$ EUR denotes the fare, $\delta=0.005$ is the congestion sensitivity parameter, $\text{VOT}=0.293/60$ EUR per minute is the value-of-time parameter, and $TT_{ij}$ is in-vehicle travel time.\footnote{Endogenous waiting time increases with total realized ridership: $T^{\text{wait}}_t 
= 0.8 \left( \sum_{i,j} \tilde{Q}_{ijt} \right)^{1/3} v^{-2/3}$,
The equilibrium is obtained via iterative updates until convergence with a tolerance of $10^{-3}$ and a maximum of 100 iterations.}

\subsubsection{TPPO vs. beachmark policies}
\label{sec:5.4.1}

To assess the versatility of TPPO in the NYC MoD service region design problem, we compare it against benchmark policies under two experimental settings. First, we remove the $k$-region constraint and the spillover effect, reducing the model to an unconstrained sequential investment problem with independent regional demand dynamics. Under this simplified structure, any subset of regions can be deployed at each stage. TPPO is then compared with an all-in deployment policy, which invests in all candidate regions at the initial decision epoch, reflecting conventional one-shot expansion without intertemporal flexibility. Second, we restore the full model specification, including both the $k$-region constraint and spillover interactions, and compare TPPO with two myopic heuristic policies under the original sequential setting.

Because option value captures the value of timing flexibility, it is not well-defined for the one-shot all-in strategy. Therefore, we evaluate policies using expected future NPV and profitability. Expected future NPV measures the total discounted payoff generated by a deployment strategy, whereas profitability normalizes discounted payoff by realized ridership, providing a scale-adjusted measure of deployment efficiency. The expected future NPV is computed as  
\begin{equation}
\mathbb{E}[\mathrm{NPV}]
=
\frac{1}{|\Omega|}
\sum_{\omega \in \Omega}
\sum_{t_n \in \mathcal{T}}
(1+\rho)^{-t_n}
\sum_{h=1}^{|\mathcal{Z}_{\mathrm{cov}}(t_n)|}
\pi_{z_h}(t_n, X_{z_h,t_n}(\omega)),
\end{equation}
where $\Omega$ denotes the set of Monte Carlo demand paths, $\rho$ is the discount rate, and $\mathcal{Z}_{\mathrm{cov}}(t_n)$ is the set of portfolios already deployed up to time $t_n$. This metric measures the absolute discounted economic value generated by a deployment policy.  
  
Profitability is calculated as the discounted payoff-to-ridership ratio aggregated across demand paths:  
\begin{equation}
PV_{\mathrm{profit}}
=
\frac{1}{|\Omega|}
\sum_{\omega \in \Omega}
\sum_{t_n \in \mathcal{T}}
(1+\rho)^{-t_n}
\frac{
\sum_{h=1}^{|\mathcal{Z}_{\mathrm{cov}}(t_n)|}
\pi_{z_h}(t_n, X_{z_h,t_n}(\omega))
}{
\sum_{h=1}^{|\mathcal{Z}_{\mathrm{cov}}(t_n)|}
X_{z_h,t_n}(\omega)
}.
\end{equation}

Table~\ref{tab:table7} reports the results. When the $k$-region constraint and spillover effects are removed, TPPO continues to outperform the all-in deployment benchmark in both NYC scenarios. In the 7-region case, TPPO increases expected future NPV by 34.7\% (6474.12 vs. 4806.26), while profitability rises from 0.51 to 2.03. Similar improvements are observed in the 8-region case, where E[NPV] increases by 28.7\% (10421.04 vs. 8099.36) and profitability improves from 0.51 to 1.92. These results indicate that even in an unconstrained environment with independent regional demand, staged deployment generates substantial economic benefits relative to one-shot expansion. When the full model specification is restored, including the $k$-region constraint and the spillover effect, TPPO remains dominant. In the 7-region scenario, TPPO achieves the highest E[NPV] (8133.27), exceeding the best myopic heuristic by 10.4\%, and attains the highest profitability (2.95). In the 8-region case, TPPO improves E[NPV] by 11.3\% over the strongest myopic heuristic (13816.40 vs. 12419.61) and maintains the highest profitability (2.93). Overall, TPPO consistently outperforms the benchmarks across both simplified and constrained settings, demonstrating its versatility.

\begin{table}[]
\resizebox{\textwidth}{!}{%
\begin{tabular}{
>{\columncolor[HTML]{FFFFFF}}c 
>{\columncolor[HTML]{FFFFFF}}c ccccc}
\hline
\cellcolor[HTML]{FFFFFF} & \cellcolor[HTML]{FFFFFF} & \multicolumn{2}{c}{\cellcolor[HTML]{FFFFFF}7-region (NYC)} & \multicolumn{3}{c}{\cellcolor[HTML]{FFFFFF}8-region (NYC)} \\ \cline{3-7} 
\multirow{-2}{*}{\cellcolor[HTML]{FFFFFF}Problem setting} & \multirow{-2}{*}{\cellcolor[HTML]{FFFFFF}Strategy} & \cellcolor[HTML]{FFFFFF}E{[}NPV{]} & \multicolumn{2}{c}{\cellcolor[HTML]{FFFFFF}Profitability} & \cellcolor[HTML]{FFFFFF}E{[}NPV{]} & \cellcolor[HTML]{FFFFFF}Profitability \\ \hline
\cellcolor[HTML]{FFFFFF} & All-in deployment & 4806.26 & \multicolumn{2}{c}{0.51} & 8099.36 & 0.51 \\
\multirow{-2}{*}{\cellcolor[HTML]{FFFFFF}No k-region constraint \& spillover effect} & TPPO & 6474.12 & \multicolumn{2}{c}{2.03} & 10421.04 & 1.92 \\ \hline
\cellcolor[HTML]{FFFFFF} & Myopia-H & 7360.76 & \multicolumn{2}{c}{2.05} & 12133.52 & \cellcolor[HTML]{FFFFFF}1.42 \\
\cellcolor[HTML]{FFFFFF} & Myopia-L & 7537.53 & \multicolumn{2}{c}{1.54} & 12419.61 & 0.88 \\
\multirow{-3}{*}{\cellcolor[HTML]{FFFFFF}With k-region constraint \& spillover effect} & TPPO & 8133.27 & \multicolumn{2}{c}{2.95} & 13816.40 & 2.93 \\ \hline
\end{tabular}%
}
\caption{Comparison of strategies (7-region \& 8-region scenarios in NYC)}
\label{tab:table7}
\end{table}

\subsubsection{Sensitivity analyses: dynamic cost and scale effect}
\label{sec:5.4.2}

In the previous experiments, we followed the settings in Rhow (2024) and assumed constant intra-region cost $c_w$ and inter-region cost $c_i$ throughout the planning horizon. In addition, potential scale effects, i.e., cost reductions associated with network expansion, were not explicitly modeled. While these assumptions facilitate comparison with the benchmark study, they may oversimplify real-world MoD deployment environments. Therefore, we introduce two extensions: (i) a terminal cost coefficient that captures dynamic cost evolution over time, and (ii) a scale sensitivity coefficient that reflects the marginal cost reduction induced by network expansion. Specifically, we allow both intra-region and inter-region costs to vary as the investment process unfolds.

First, to model time-dependent cost dynamics, we define a time-scaling function
\begin{equation}
f_{\text{time}}(t_n) = (f_{\text{end}})^{t_n/T},
\end{equation}
where $f_{\text{end}}$ denotes the terminal cost coefficient at the end of the planning horizon $T$. When $f_{\text{end}} < 1$, costs decline over time (e.g., due to technological improvement or operational learning); when $f_{\text{end}} > 1$, costs increase (e.g., due to input price inflation or regulatory pressures). The dynamic intra-region cost is therefore given by
\begin{equation}
c_{intra}(t_n) = c_{intra} \cdot f_{\text{time}}(t_n),
\end{equation}

Second, to incorporate the scale effect, we assume that inter-region cost decreases with the size of the deployed service network. Let $|Z_{cov}(t_n)|$ denote the number of covered regions at time $t_n$. The inter-region cost is modeled as
\begin{equation}
c_{inter}(t_n) = c_{inter} \cdot f_{\text{time}}(t_n) \cdot \frac{1}{1 + \zeta |Z_{cov}(t_n)|}, 
\quad \zeta > 0,
\end{equation}
where $\zeta$ is the scale sensitivity coefficient. A larger $\zeta$ implies stronger marginal cost reduction as the service network expands.

\begin{figure}[htbp]
    \centering
    \includegraphics[width=0.6\textwidth]{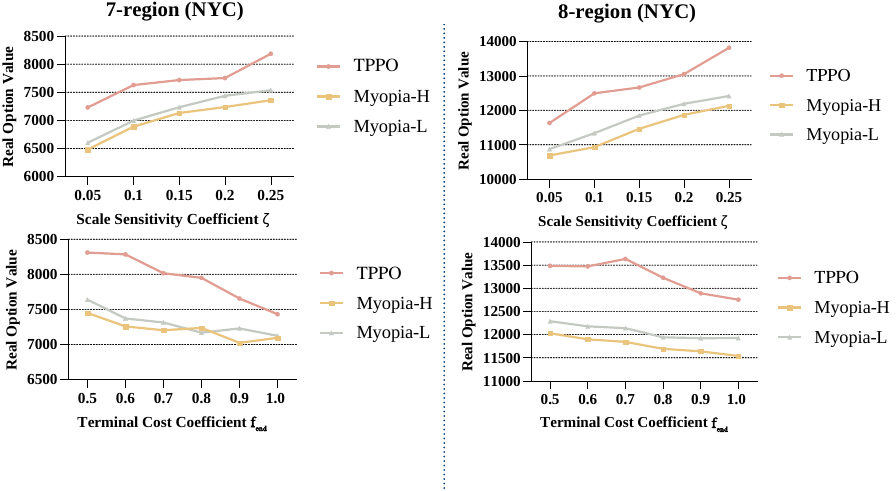}
    \caption{Sensitivity analyses of dynamic cost\& scale effect (7-region \& 8-region scenarios in NYC)}
    \label{fig:fig7}
\end{figure}

Figure~\ref{fig:fig7} compares the real option values of investment sequences generated by TPPO and the two myopic heuristic policies under varying levels of the scale sensitivity coefficient $\zeta$ and the terminal cost coefficient $f_{\text{end}}$. Across both the 7-region and 8-region NYC scenarios, TPPO consistently achieves the highest option value under all parameter settings. As  $\zeta$ increases, the option value rises monotonically for all policies, reflecting stronger economies of scale from network expansion. However, TPPO exhibits a steeper improvement in option value as $\zeta$ increases, indicating that the learning-based policy better internalizes scale-dependent cost dynamics when constructing investment sequences. By contrast, myopic heuristics are less capable of leveraging such intertemporal cost advantages. The impact of $f_{\text{end}}$ differs across policies. When $f_{\text{end}}$ is lower (i.e., costs decline more substantially over time), the value of postponement increases because future investments can be executed at lower cost. Under such settings, TPPO achieves a larger performance margin over the myopic heuristics. This suggests that TPPO can make use of the future cost reductions and strategically delays selected investments to exploit higher option value, whereas the myopic policies lack the agility and underutilize the intertemporal cost advantage. As $f_{\text{end}}$ increases and the benefit of waiting diminishes, the relative advantage of TPPO correspondingly narrows, although it remains dominant across all settings.

\section{Conclusion}
\label{sec:6}

This study considers a sequential service region design (SSRD) that determines the timing and ordering of regional expansion to maximize long-term service network revenue. Because full-scale simultaneous deployment is typically infeasible under capital and operational constraints, expansion must proceed in stages. The core challenge is therefore to determine when to invest and which regions to deploy at each stage. This decision is complicated by demand uncertainty, intertemporal trade-offs between early and delayed investment, and network effects whereby each deployment reshapes future demand through inter-regional connectivity. To capture these complexities, we formulate SSRD as an MDP under a real options framework over a finite planning horizon, where heterogeneous regions exhibit nonstationary stochastic demand composed of intra- and inter-region flows. Extending \textcite{Rath_2024_deepa_S16}, we incorporate a $k$-region constraint that limits the number of regions investable per period and introduce a stochastic spillover effect linking investment decisions to demand evolution. The resulting problem requires sequencing regional portfolios under uncertainty, leading to a combinatorial explosion in feasible investment sequences. To address this computational challenge, we integrate real options analysis (ROA) with a Transformer-based Proximal Policy Optimization (TPPO) framework within the MDP formulation. ROA evaluates the intertemporal option value of complete investment trajectories under uncertainty, while TPPO learns sequential investment policies whose rollout generates high option-value expansion sequences.

Extensive numerical experiments based on realistic multi-region settings in Shanghai and Beijing demonstrate the effectiveness of the proposed framework. TPPO consistently converges faster than benchmark DRL methods and identifies investment sequences with higher option value. In small-scale instances where full enumeration is feasible, TPPO achieves near-optimal option value solutions with an average optimality gap of only 1.31\% while substantially reducing computational time. In larger networks, it continues to outperform myopic heuristics by discovering sequences that capture greater option value. A mobility-on-demand service area expansion case study in New York City further validates the practical applicability of the approach. Under both unconstrained and fully constrained settings, TPPO consistently surpasses one-shot deployment and myopic policies, significantly enhancing expected future NPV and profitability. Sensitivity analyses across diverse spillover structures, dynamic cost trajectories, and scale effects confirm the robustness of TPPO under varied assumptions.

Beyond numerical performance, this study provides several theoretical and managerial contributions. From a theoretical perspective, we extend the SSRD literature along three dimensions. First, by introducing the $k$-region constraint, we shift the problem from sequencing individual regions to sequencing feasible regional portfolios, thereby reflecting realistic capital and operational limits while substantially increasing decision complexity. Second, by incorporating the stochastic spillover effect linked to real-time investment decisions, we enrich the modeling of nonstationary demand dynamics and capture endogenous network externalities that have been underexplored in prior SSRD studies. Third, we integrate ROA with a DRL framework to address the resulting combinatorial challenge. This integration demonstrates how learning-based policies can internalize intertemporal investment flexibility and cross-regional dependencies, offering a scalable approach for various SSRD scenarios.

From a managerial perspective, several insights emerge. First, optimal investment sequences are structurally stable across different concurrency limits ($k$). Increasing $k$ rarely changes the investment order or requires fully exhausting the constraint. Although relaxing the $k$-region limit enhances flexibility, its value is non-monotonic and typically peaks at moderate levels (e.g., $k=4$–$5$ in the 9-region case). Concurrency effectiveness depends on demand conditions: conservative pacing (small $k$) suffices in stable markets, moderate concurrency creates greater value under growth or volatility, whereas fully aggressive rollout (large $k$) is rarely optimal. Second, the TPPO-derived sequences follow a clear bottom-up logic: regions with lower baseline demand and smaller scale are deployed earlier to capture quick returns, while high-demand regions are strategically deferred. Third, investment concurrency is selective rather than uniform. High-value sequences reveal that certain region pairs are consistently co-invested, indicating structured complementarities. Finally, as spillover intensity increases, whether positive or negative, the advantage of our DRL-based policy over myopic policies becomes more pronounced, since myopic policies fail to adapt to evolving demand. A similar amplification effect arises when the scale effect strengthens or service costs decline over time.

Several research directions emerge from this work. 
First, the framework could incorporate richer investment constraints, such as regional precedence requirements or performance-triggered expansion thresholds, to enhance practical relevance. Second, the model could allow for greater investment flexibility, including temporary suspension, abandonment, or reinvestment of previously deployed regions. Embedding such dynamic adjustment options would further strengthen the connection between SSRD and real options theory. Third, future research could integrate customer-centric performance metrics, such as service accessibility and equity, to examine trade-offs between investment option value and broader social welfare objectives. Finally, while this study assumes exogenous demand evolution driven by the spillover effect, incorporating endogenous competitive market responses would provide a richer representation of platform dynamics.

% \section{CRediT authorship contribution statement}
% Tingting Chen: Writing – review & editing, Methodology, Formal analysis, Conceptualization. İzzet Egemen Elver: Writing – original draft, Methodology, Formal analysis. Tuna Arda Kınık: Writing review & editing, Software.

\section{Declaration of competing interest}
The authors declare that they have no known competing financial interests or personal relationships that could have appeared to influence the work reported in this paper. 

\section{Data availability}
Data will be made available on request.

% \bibliography{ref}
\printbibliography

\appendix
\section{Appendix}
\label{sec:7}

\begin{table}[htbp]
\centering
% --- 修改 1: 将表格编号重命名为 A1 ---
\renewcommand{\thetable}{A1}

% --- 修改 2: 调整宽度 ---
% 将 \textwidth 改为 0.65\textwidth (或者您想要的任何比例，如 0.8)
\resizebox{0.65\textwidth}{!}{%
    % --- 修改 3: 移除了多余的空列 (原来的 lllll) ---
    \begin{tabular}{@{}
    >{\columncolor[HTML]{FFFFFF}}c
    >{\columncolor[HTML]{FFFFFF}}c
    >{\columncolor[HTML]{FFFFFF}}c
    >{\columncolor[HTML]{FFFFFF}}c
    >{\columncolor[HTML]{FFFFFF}}c @{}}
    \toprule
    % 将原来的第一行文字移动到了 caption 中，这里保留表头
    City name & Region index & Region name & Area (km²) & Population density (/km²) \\ \midrule
    \cellcolor[HTML]{FFFFFF} & r1 & Changning & 37.16 & 18164 \\
    \cellcolor[HTML]{FFFFFF} & r2 & Xuhui & 55.16 & 20358 \\
    \cellcolor[HTML]{FFFFFF} & r3 & Jingan & 36.77 & 26193 \\
    \cellcolor[HTML]{FFFFFF} & r4 & Huangpu & 20.5 & 28451 \\
    \cellcolor[HTML]{FFFFFF} & r5 & Yangpu & 60.55 & 30443 \\
    \cellcolor[HTML]{FFFFFF} & r6 & Hongkou & 23.41 & 22069 \\
    \cellcolor[HTML]{FFFFFF} & r7 & Putuo & 36.77 & 22681 \\
    \multirow{-8}{*}{\cellcolor[HTML]{FFFFFF}Shanghai} & r8 & Minhang & 373.00 & 7210 \\ \midrule
    \cellcolor[HTML]{FFFFFF} & r1 & Dongcheng & 41.79 & 16962 \\
    \cellcolor[HTML]{FFFFFF} & r2 & Xicheng & 50.47 & 21918 \\
    \cellcolor[HTML]{FFFFFF} & r3 & Shijingshan & 85.05 & 6677 \\
    \cellcolor[HTML]{FFFFFF} & r4 & Haidian & 431.00 & 7270 \\
    \cellcolor[HTML]{FFFFFF} & r5 & Chaoyang & 464.90 & 7426 \\
    \cellcolor[HTML]{FFFFFF} & r6 & Fengtai & 304.90 & 6624 \\
    \cellcolor[HTML]{FFFFFF} & r7 & Changping & 1342.00 & 1691 \\
    \cellcolor[HTML]{FFFFFF} & r8 & Shunyi & 1009.00 & 1312 \\
    \multirow{-9}{*}{\cellcolor[HTML]{FFFFFF}Beijing} & r9 & Tongzhou & 888.80 & 2071 \\ \bottomrule
    \end{tabular}%
}
% --- 添加 Caption 以显示 Table A1 ---
\caption{Region attributes in Shanghai \& Beijing, China}
\label{tab:tableA1}
\end{table}

\end{document}